%% file: main.tex
\documentclass[journal]{IEEEtran}

\ifCLASSINFOpdf
\else
\fi

\usepackage{amsmath}
\usepackage{amsthm}
\usepackage{times}
\usepackage{graphicx,subcaption}
\usepackage{color}
\usepackage{multirow}
\usepackage{enumitem}
\usepackage{cite}
\usepackage{amsfonts}
\usepackage{rotating}
\usepackage{bbm}
\usepackage{latexsym}
\usepackage[table,xcdraw]{xcolor}
\usepackage{colortbl}
\usepackage{mathtools}
\usepackage{mathrsfs}

\usepackage{capt-of}
\usepackage{minipage}

\theoremstyle{definition}
\newtheorem{definition}{Definition}
\newtheorem{prop}{Proposition}
\newtheorem{thm}{Theorem}

\begin{document}

\title{A Comparative Study of Rule Extraction for Recurrent Neural Networks}

\author{Qinglong Wang,~\IEEEmembership{Student Member, IEEE},
        Kaixuan Zhang,
        Alexander G. Ororbia II, \\
        Xinyu Xing, 
        Xue Liu,~\IEEEmembership{Member,~IEEE}, 
        C. Lee Giles,~\IEEEmembership{Fellow,~IEEE}
\thanks{Q. Wang and X. Liu are with the School of Computer Science at McGill University.}
\thanks{K. Zhang, A. Ororbia II, X. Xing and C.L. Giles are with at the College of Information Sciences and Technology at the Pennsylvania State University.}}

%
%


\maketitle

\input{base/abstract}

\begin{IEEEkeywords}
Deterministic finite automata, recurrent neural networks, rule extraction, grammar complexity.
\end{IEEEkeywords}

%
\IEEEpeerreviewmaketitle

\input{base/1_0_intro}

\input{base/2_0_background} 
\input{base/3_0_formulation}
\input{base/4_0_complexity}

\input{base/4_1_entropy}

\input{base/4_2_distance}

\input{base/4_3_entropy_dist}

\input{base/5_0_exp_config}
\input{base/5_1_acc}

\input{base/5_2_success}

\input{base/6_0_conclusion}



\bibliographystyle{IEEEtran}
\bibliography{ref}

\ifCLASSOPTIONcaptionsoff
  \newpage
\fi

\input{base/appendix}



%

\end{document}

%% file: base/abstract.tex
\begin{abstract}
Understanding recurrent networks through rule extraction has a long history. This has taken on new interests due to the need for interpreting or verifying neural networks. 
One basic form for representing stateful rules is deterministic finite automata (DFA). Previous research shows that extracting DFAs from trained second-order recurrent networks is not only possible but also relatively stable. Recently, several new types of recurrent networks with more complicated architectures have been introduced. These handle challenging learning tasks usually involving sequential data. However, it remains an open problem whether DFAs can be adequately extracted from these models. Specifically, it is not clear how DFA extraction will be affected when applied to different recurrent networks trained on data sets with different levels of complexity. 
Here, we investigate DFA extraction on several widely adopted recurrent networks that are trained to learn a set of seven regular Tomita grammars. We first formally analyze the complexity of Tomita grammars and categorize these grammars according to that complexity. Then we empirically evaluate different recurrent networks for their performance of DFA extraction on all Tomita grammars. Our experiments show that for most recurrent networks, their extraction performance decreases as the complexity of the underlying grammar increases. On grammars of lower complexity, most recurrent networks obtain desirable extraction performance. As for grammars with the highest level of complexity, while several complicated models fail with only certain recurrent networks having satisfactory extraction performance. 
\end{abstract}

%% file: base/1_0_intro.tex
\section{Introduction}
\label{sec:intro}
Recurrent neural networks (RNNs) are one of the most powerful learning models for processing sequential data and, like neural networks in general, have often been considered as ``black-box'' models. This black-box nature is largely due to the fact that RNNs, as much as any neural architecture, are designed to capture structural information from the data and store learned knowledge in the synaptic connections between nodes (or weights)~\cite{du2017topology}. This makes inspection, analysis, and verification of captured knowledge difficult or near-impossible~\cite{omlin2000symbolic}. One approach to this problem is to investigate if and how we might extract symbolic knowledge from trained RNNs, since symbolic knowledge is usually regarded as easier to understand. Surprisingly, this is an old problem that was treated by Minsky in the chapter titled ``Neural Networks. Automata Made up of Parts'' in his text ``Computation, Finite and Infinite Machines''~\cite{minsky1967computation}. Specifically, if one treats the information processing of a RNN as a mechanism for representing knowledge in symbolic form where a set of rules that govern transitions between symbolic representations are learned, then the RNN can be viewed as an automated reasoning process with production rules, which should be easier to understand. 

Prior work focused on extracting symbolic knowledge from recurrent networks. As an example, Borges et al.~\cite{BorgesGL11} proposed to extract symbolic knowledge from a nonlinear autoregressive model with exogenous inputs model (NARX)~\cite{Lin96NARX}. Also, it has been demonstrated that by representing information of long-term dependencies in the form of symbolic knowledge~\cite{dhingra2017linguistic}, RNNs' ability for handling long-term dependencies can be improved. In sentiment analysis, recent work~\cite{murdoch2017automatic} shows that recurrent networks can be explained by decomposing their decision making process and identifying patterns of words which are ``believed'' to be important for the decision making. If the words are viewed as symbols, then their patterns can be regarded as representing the rules for determining that sentiment. One of the most frequently adopted rule extraction approaches is to extract deterministic finite automata (DFA) from recurrent networks trained to perform grammatical inference~\cite{Empirical2017Wang,weiss2017extracting,cohen2017inducing,omlin1996extraction, casey1996dynamics,giles1992learning,watrous1992induction}. Approaches following this direction are categorized as ~\emph{compositional}~\cite{jacobsson2005rule}. In particular, the vector space of a RNN's hidden layer is first partitioned into finite elements, where each part is treated as a state of a certain DFA. Then, transitions rules that are associated with the alphabet at that time connecting these states are extracted (also known as production rules). Using a DFA to represent production rules is motivated by the need for conducting comprehensive analysis and understanding of the computational abilities of recurrent networks~\cite{jacobsson2005rule}. 

Recent work~\cite{Empirical2017Wang} demonstrates that extracting DFAs from a second-order RNN~\cite{giles1991second} is not only possible, but the extraction is relatively stable even when the hidden layer of a second-order RNN is randomly initialized. The latter is important because it has been argued that when a RNN is viewed as a dynamical system, its training process is too sensitive to the initial state of the model~\cite{kolen1994fool}. This implied that the following DFA extraction may be unstable which has been shown to be not the case. 

Despite much prior work on rule extraction from recurrent networks including Elman networks (Elman-RNN)~\cite{elman1990finding} and second-order RNNs~\cite{giles1992learning,watrous1992induction}, little is known if the aforementioned compositional approaches can be effectively applied to other recurrent networks, especially those that have demonstrated impressive performance on various sequence learning tasks, e.g. long-short-term-memory networks (LSTM)~\cite{hochreiter1997long}, gated-recurrent-unit networks (GRU)~\cite{cho2014properties}, multiplicative integration recurrent neuron networks (MI-RNN)~\cite{wu2016multiplicative}, etc. Another equally important yet missing study is how and if DFA extraction will be affected by the data source on which recurrent networks are trained.

In this work, we greatly expand upon previous work~\cite{giles1992learning, Empirical2017Wang} and study the effect of DFA extraction performance on different types of recurrent networks and data sets with different levels of complexity. Specifically, the recurrent networks investigated in this study include Elman-RNN, second-order RNN, MI-RNN, LSTM and GRU. We follow previous work~\cite{Empirical2017Wang,weiss2017extracting,li2016kernel,casey1996dynamics,omlin1996extraction,giles1992learning,watrous1992induction} by adopting a family of seven relatively simple, yet important, regular grammars, which were originally proposed by Tomita~\cite{tomita1982} and widely studied and used as benchmarks for DFA extraction. Given a recurrent model and a Tomita grammar, the performance is evaluated by measuring both the quality of DFAs extracted from this model, and the success rate of extracting the correct DFA which is identical to the unique DFA associated with the grammar used for training that model. Both metrics are evaluated for multiple random trails in order to evaluate the overall performance of DFA extraction for all recurrent networks on all Tomita grammars. In summary, we make the following contributions:  
\begin{itemize}[leftmargin=*]
    \item We analyze and categorize regular grammars with a binary alphabet (including Tomita grammars) by defining a entropy that describes their complexity. We discuss the difference between our defined entropy and the entropy of shift space defined for describing symbolic dynamics, and show that our definition is more informative for describing the complexity of regular grammars and can be extended to multiclass classification problems.
    \item We propose an alternative metric -- the averaged edit distance -- for describing the complexity of regular grammars with a binary alphabet. We show that this metric is closely related to our defined entropy of regular grammars. In addition, through experiments, we demonstrate that the average edit distance reflects a more defining complexity of Tomita grammars and our defined entropy is more computationally efficient to calculate.
    \item We conduct a careful experimental study of evaluating and comparing different recurrent networks for DFA extraction. Our results show that among all RNNs investigated, RNNs with quadratic (or approximate quadratic) forms of hidden layer interaction, i.e. second-order RNN and MI-RNN, provide the most accurate and stable DFA extraction for all Tomita grammars. In particular, on grammars with a high level of complexity, second-order RNN and MI-RNN achieve much better success rates of DFA extraction than other recurrent networks.
\end{itemize}

%% file: base/2_0_background.tex
\section{Background}
\label{sec:background}

In this section, we first briefly introduce DFAs and regular grammars, followed by the set of Tomita grammars used in this study. Then we introduce existing rule extraction methods, especially the compositional approaches which are most widely studied for extracting DFA from recurrent networks.

\subsection{Deterministic Finite Automata}
\label{sec:dfa}
Based on the Chomsky hierarchy of phrase structured grammars~\cite{chomsky1956three}, a regular grammar is associated with one of the simplest automata, a deterministic finite automata (DFA). Specifically, given a regular grammar $G$, it can be recognized and generated by a DFA $M$, which can be described by a five-tuple $\{\Sigma, S, s_0, s_F, P\}$. $\Sigma$ is the input alphabet (a finite, non-empty set of symbols) and $S$ is a finite, non-empty set of states. $s_0 \in S$ represents the initial state while $s_F \in S$ represents the set of final states.\footnote{Note that $s_F$ can be the empty set, $\emptyset$.} $P$ denotes a set of deterministic production rules. Every grammar $G$ also recognizes and generates a corresponding language, a set of strings of symbols from alphabet $\Sigma$. It is important to realize that DFA covers a wide range of languages which means that all languages whose string length and alphabet size are bounded can be recognized and generated by a DFA~\cite{giles1992learning}. Also, when replacing the deterministic transition with stochastic transition, a DFA can be converted as a probabilistic automata or hidden Markov model, which enables the use of graphical models~\cite{du2016convergence} for grammatical inference. For a more thorough and detailed treatment of regular language and finite state machines, please refer to~\cite{hopcroft2006automata}.

\subsection{Tomita Grammars}
\label{sec:tomita}
Tomita grammars~\cite{tomita1982} denote a set of seven regular grammars that have been widely adopted in the study of extracting DFA from RNNs. In principle, when compared with regular grammars associated with large finite-state automata, Tomita grammars should be easily learnable, given that the DFAs associated with Tomita grammars have between three and six states. These grammars all have alphabet $\Sigma = \{0,1\}$, and generate an infinite language over $\{0,1\}^{*}$. For each Tomita grammar, we refer to the binary strings generated by this grammar as its associated positive examples and other binary strings as negative examples. A description of Tomita grammars is provided in Table~\ref{tab:tomita} and the DFA for grammar 2 is shown as an example in Figure~\ref{fig:g2_dfa}. 

\begin{table*}[t]
\begin{minipage}[b]{0.6\linewidth}
\centering
\begin{tabular}{clcccc}
  \hline \hline
  G & Description                                                             & Training     & Testing     & Length     & Parameters\\ \hline \hline
  1 & $1^{*}$                                                                 & 7.8\%    & 92.2\%   & 1-14   & 1220    \\ \hline
  2 & $(1 0)^{*}$                                                             & 6.5\%    & 93.5\%   & 2-14  & 1220     \\ \hline
  3 & \begin{tabular}[c]{@{}l@{}}an odd number of consecutive 1s is\\
       always followed by an even number\\of consecutive 0s
      \end{tabular}                                                           & 36.7\%       & 63.3\%      & 4-12  & 1220     \\ \hline
  4 & \begin{tabular}[c]{@{}l@{}}any string not containing ``000'' as a\\
      substring \end{tabular}                        & 36.7\%       & 63.3\%      & 3-12  & 1220     \\ \hline
  5 & \begin{tabular}[c]{@{}l@{}}even number of 0s and even number\\
      of 1s~\cite{giles1990higher}  \end{tabular}        & 36.7\%       & 63.3\%      & 4-12  & 30100     \\ \hline
  6 & \begin{tabular}[c]{@{}l@{}}the difference between the number of 0s\\
      and the number of 1s is a multiple of 3
      \end{tabular}                                                           & 36.7\%       & 63.3\%      & 3-12  & 10502     \\ \hline
  7 & $0^{*}1^{*}0^{*}1^{*}$                                                  & 8.9\%    & 81.7\%   & 1-16  & 1220     \\ \hline \hline
  \end{tabular}
    \caption{Descriptions of the Tomita grammars and the configuration of their data sets. The number of parameters specified for each recurrent networks are either same or closest to the values shown in the ``Paramters'' column.}
    \label{tab:tomita}
\end{minipage}\hfill
\begin{minipage}[b]{0.35\linewidth}
\centering
\includegraphics[width=0.55\linewidth]{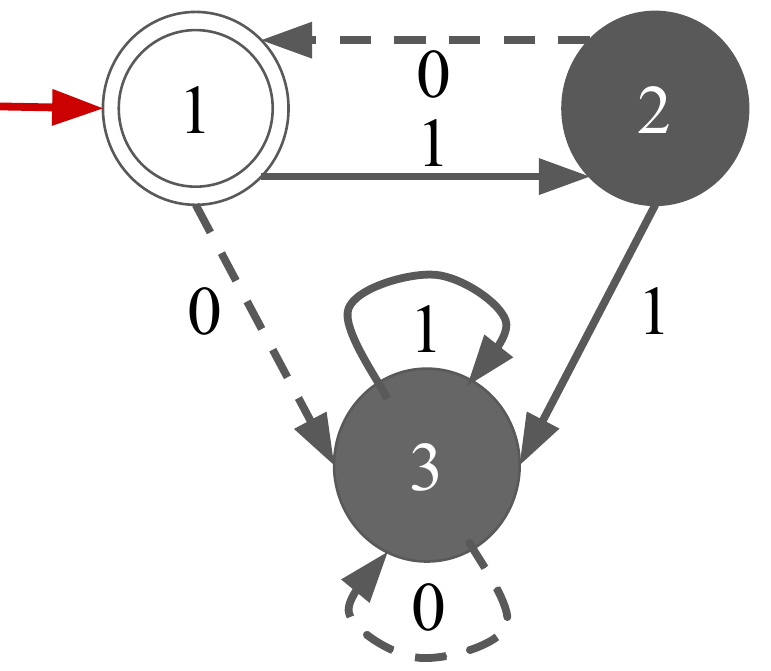}
\captionof{figure}{Example DFA for Tomita grammar 2. Red arrow indicates the initial state, shaded circles indicate non-accept states. Dotted lines indicate input ``0'' and solid lines indicate input ``1''.}
\label{fig:g2_dfa}
\end{minipage}
\end{table*}

Despite being relatively simple, Tomita grammars actually represent regular grammars with a wide range of complexity. As shown in Table~\ref{tab:tomita}, the distinction between positive and negative samples for different grammars are very different. For instance, grammars 1, 2 and 7 represent the class of regular languages that define a string set that has extremely unbalanced positive and negative strings. This could represent real-world cases where positive samples are significantly outnumbered by negative ones. In contrast, grammars 5 and 6 define the class of regular languages that have equal or a relatively balanced number of positive and negative strings. In particular, on grammar 5, the numbers of positive and negative strings are the same for string sets with even length. This indicates that the difference between positive and negative strings in these grammars is much smaller than the case for grammars 1,2 and 7. The difference between the numbers of positive and negative strings for grammar 3 an 4 lies between the above cases. When constructing RNNs to learn Tomita grammars, with either case discussed, RNNs are forced to recognize the various levels of difference between positive and negative samples. 

It is worth mentioning that the popularity of Tomita grammars in studying DFA extraction problem is also due to the fact that the ground truth DFAs for Tomita grammars are available. This enables previous studies~\cite{giles1992learning, Empirical2017Wang} to determine the impact of different factors on the performance of DFA extraction by comparing extracted DFAs with ground truth DFAs. More complex/or real-world data sets may not well support preliminary studies of DFA extraction, since for those datasets, uncertainties will be introduced into the evaluation (e.g. what is the ground truth DFA or if there even exists ground truth DFAs that define the data?). We believe this uncertainty can affect any conclusion of whether a DFA extraction can be stably performed.

\subsection{Rule Extraction for recurrent networks}
\label{sec:rule_extraction}
A survey~\cite{jacobsson2005rule} on rule extraction methods for recurrent networks categorizes them as (1)~compositional approaches, which categorize the cases when rules are constructed based on the hidden layers -- ensembles of hidden neurons -- of a RNN; (2)~decompositional approaches, where rules are constructed based on individual neurons; (3)~pedagogical approaches, which construct rules by regarding the target RNN as a black box and have no access to the inner state of this RNN; and (4)~eclectic approaches, which represent a hybrid of decompositional and pedagogical approaches. Most aforementioned approaches conduct rule extraction in a post hoc manner. That is, rule extraction is performed with an already trained RNN and a data set containing samples to be processed by this RNN.

Using DFA as rules extracted from RNNs~\cite{giles1992learning, omlin1996extraction, casey1996dynamics, Empirical2017Wang, weiss2017extracting} as been very common. In these studies, a RNN is viewed as representing the state transition diagram of a state process -- $\{$input, state$\} \Rightarrow \{$next state$\}$ -- of a DFA. Correspondingly, a DFA extracted from a RNN can globally describe the behavior of this RNN. Recent work~\cite{murdoch2017automatic, Murdoch18Beyond, Murdoch18Hierarchical} proposes to extract instance-based rules. Specifically, individual rules are extracted from data instances and each extracted rule represents a pattern. As shown in the case for sentiment analysis, a pattern is a combination of important words identified from a sentence processed by a RNN to be interpreted. To construct a global rule set that describes the most important patterns learned by the target RNN, extracted individual rules need to be aggregated using statistical methods~\cite{murdoch2017automatic}. A rule set constructed in this manner usually lacks formal representation and may not be suitable for conducting a more thorough analysis of the behaviors of a RNN. In this work, we follow previous work~\cite{giles1992learning, omlin1996extraction, casey1996dynamics, Empirical2017Wang, weiss2017extracting} and represent rules by DFA.

Among all of the above mentioned approaches, both pedagogical and compositional approaches have been applied to extract DFAs from RNNs. In the former category, a recent work~\cite{pmlr-v80-weiss18a} proposes to build a DFA by only querying the outputs of a RNN for certain inputs. This method can be effectively applied to regular languages with small sizes of alphabet, however, it cannot scale to languages with a large size alphabet. This is mainly due to the fact this method replies on the $\mathrm{L}^{*}$ algorithm~\cite{Angluin87} which has polynomial complexity. As a result, the extraction process becomes extremely slow when a target RNN performs complicated analysis when processing sophisticated data~\cite{pmlr-v80-weiss18a}. The compositional approaches are much more commonly adopted in previous studies~\cite{giles1992learning,gori1998inductive,jacobsson2005rule,Empirical2017Wang}. In these works, it is commonly assumed that the vector space of a RNN's hidden layer can be approximated by a finite set of discrete states~\cite{jacobsson2005rule}, where each rule refers to the transitions between states. As such, a generic compositional approach can be described by the following basic steps:
\begin{enumerate}[wide, labelwidth=!, labelindent=0pt]
\item Collect the values of a RNN's hidden layers when processing every sequence at every time step. Then quantize the collected hidden values into different states.
\item Use the quantized states and the alphabet-labeled arcs that connect these states to construct a transition diagram.
\item Reduce the diagram to a minimal representation of state transitions.
\end{enumerate}
Previous research has mostly focused on improving the quantization step~\footnote{For a more detailed discussion the other two steps, please refer to our previous work~\cite{Empirical2017Wang} and a survey~\cite{jacobsson2005rule}.}. The efficacy of different quantization methods relies on the following hypothesis. The state space of a RNN, which is well trained to learn a regular grammar, should already be fairly well separated with distinct regions that represent the corresponding states in some DFA. This hypothesis, if true, implies that much less effort is required for the quantization step. Indeed, various quantization approaches including equipartition-based methods~\cite{giles1992learning,omlin1996extraction} and clustering methods~\cite{zeng1993learning,Empirical2017Wang} have been adopted and demonstrated that this hypothesis holds for second-order recurrent networks.

%% file: base/3_0_formulation.tex
\section{Problem Formulation}
\label{sec:formulation}
Different rule extraction approaches proposed in previous works introduced in Section~\ref{sec:background} essentially describes the process of developing or finding a rule that approximates the behaviors of a target RNN~\cite{jacobsson2005rule}. In the following, we generalize the rule extraction problem in a formal manner. 

\begin{definition}[Rule Extraction Problem]
\label{def:rule_extraction}
Given a RNN denoted as a function $f: \mathcal{X} \rightarrow \mathcal{Y}$ where $\mathcal{X}$ is the data space, $\mathcal{Y}$ is the target space, and a data set $B = \{X, Y\}$ with $n$ samples $X \in \mathcal{X}^n$ and $Y \in \mathcal{Y}^{n}$. Let $r$ denote a rule which is also a function with its data and target space identical to that of $f$. The rule extraction problem is to find a function $\mathscr{L} : (\mathcal{X} \rightarrow \mathcal{Y}) \times (\mathcal{X}^n \times \mathcal{Y}^n) \rightarrow (\mathcal{X} \rightarrow \mathcal{Y}) $ such that $\mathscr{L}$ takes as input a $f$ and a $B$ then outputs a rule $r$. 
\end{definition}

As introduced in Section~\ref{sec:intro}, in this study we aim at investigating if and how will DFA extraction be affected when we apply DFA extraction to different recurrent networks trained on data sets with different levels of complexity. More specifically, in our case where the underlying data sets are generated by Tomita grammars, we denote by $B(G)$ a data set generated by a grammar $G$. Also, $\mathcal{X} = \Sigma^{\ast}$ where $\Sigma = \{0,1\}$, $\mathcal{Y} = \{0, 1\}$ represents the space of labels for positive and negative strings recognized by $G$. Then in our evaluation framework, we fix the extraction method $\mathscr{L}$ as the compositional approach (introduced in Section~\ref{sec:rule_extraction}) and evaluate the performance obtained by $\mathscr{L}$ when its input, i.e. $B(G)$ and $f$ trained on $B_{train}(G)$~\footnote{Data set $B(G)$ is split into a training set $B_{train}(G)$ and a test set $B_{test}(G)$ as typically done for supervised learning.}, vary across different grammars and different recurrent networks respectively. It is important to note that, by comparing the extraction performance obtained by a given model across different grammars, we then examine how sensitive is each model with respect to the underlying data for DFA extraction problem.

According to above definition, it is clear that the performance of DFA extraction can be evaluated by measuring the quality of extracted DFAs. To be more specific, the extraction performance is evaluated by two metrics. The first metric is the accuracy of an extracted DFA when it is tested on the test set for a certain grammar. The second metric is the success rate of extracting DFAs that are identical to the ground truth DFA associated with a certain grammar, hence should perform perfectly on the test set generated by this grammar. In other words, given a grammar, the success rate of a recurrent model measures how frequently can the correct DFA associated with this grammar be extracted. These metrics quantitatively measure the abilities of different recurrent networks for learning different grammars. In particular, the first metric reflects the abilities of different recurrent networks for learning ``good'' DFAs. Due to its generality, the first metric is also frequently adopted in much research work~\cite{Ribeiro0G16, Hinton17Distilling, murdoch2017automatic, weiss2017extracting}. The second metric, which is more rigorous in comparison with the first metric, reflects the abilities of these models for learning correct DFAs. It is important to note that our evaluation framework is agnostic to the underlying extraction method since we imposes no constraint on $\mathscr{L}$. As such, this evaluation framework can also be adopted for comparing different rule extraction methods and will be included in our future work.


%% file: base/4_0_complexity.tex
\section{The Complexity of Tomita Grammars}
\label{sec:complexity}
In this section, we analyze the complexity of Tomita grammars by defining two metrics -- the entropy and average edit distance for regular grammars. In principle, these metrics are defined to to measure how balanced are the sets of positive and negative strings, and the difference between these sets. Accordingly, a grammar with higher complexity has more balanced string sets and less difference between these sets.

\begin{figure}[!t]
\centering
\begin{subfigure}{.15\textwidth}
  \centering
  \includegraphics[width=\linewidth]{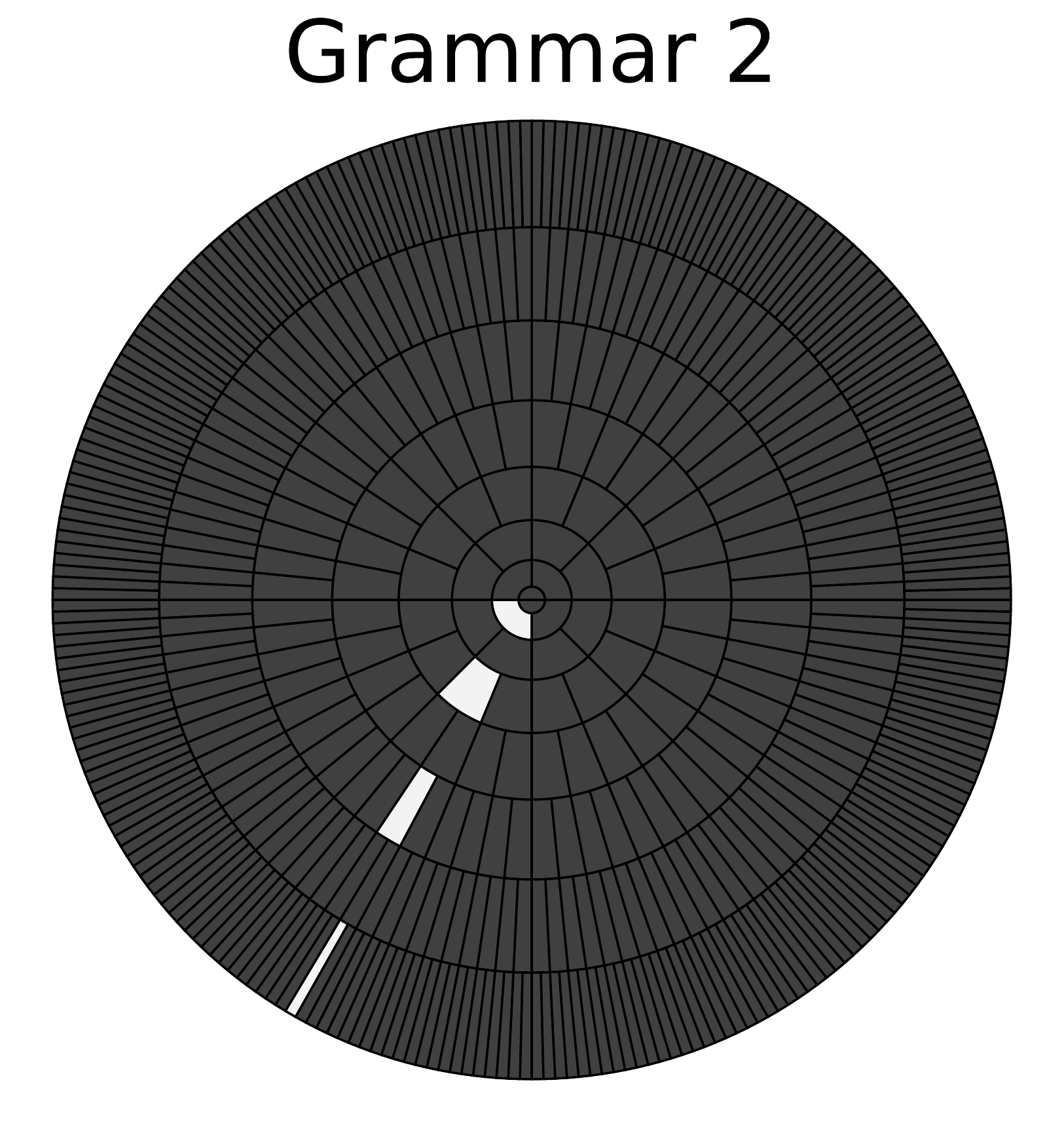}
  \vspace{-0.5em}
  \label{fig:g2_pie}
\end{subfigure} \hfill 
\begin{subfigure}{.15\textwidth}
  \centering
  \includegraphics[width=\linewidth]{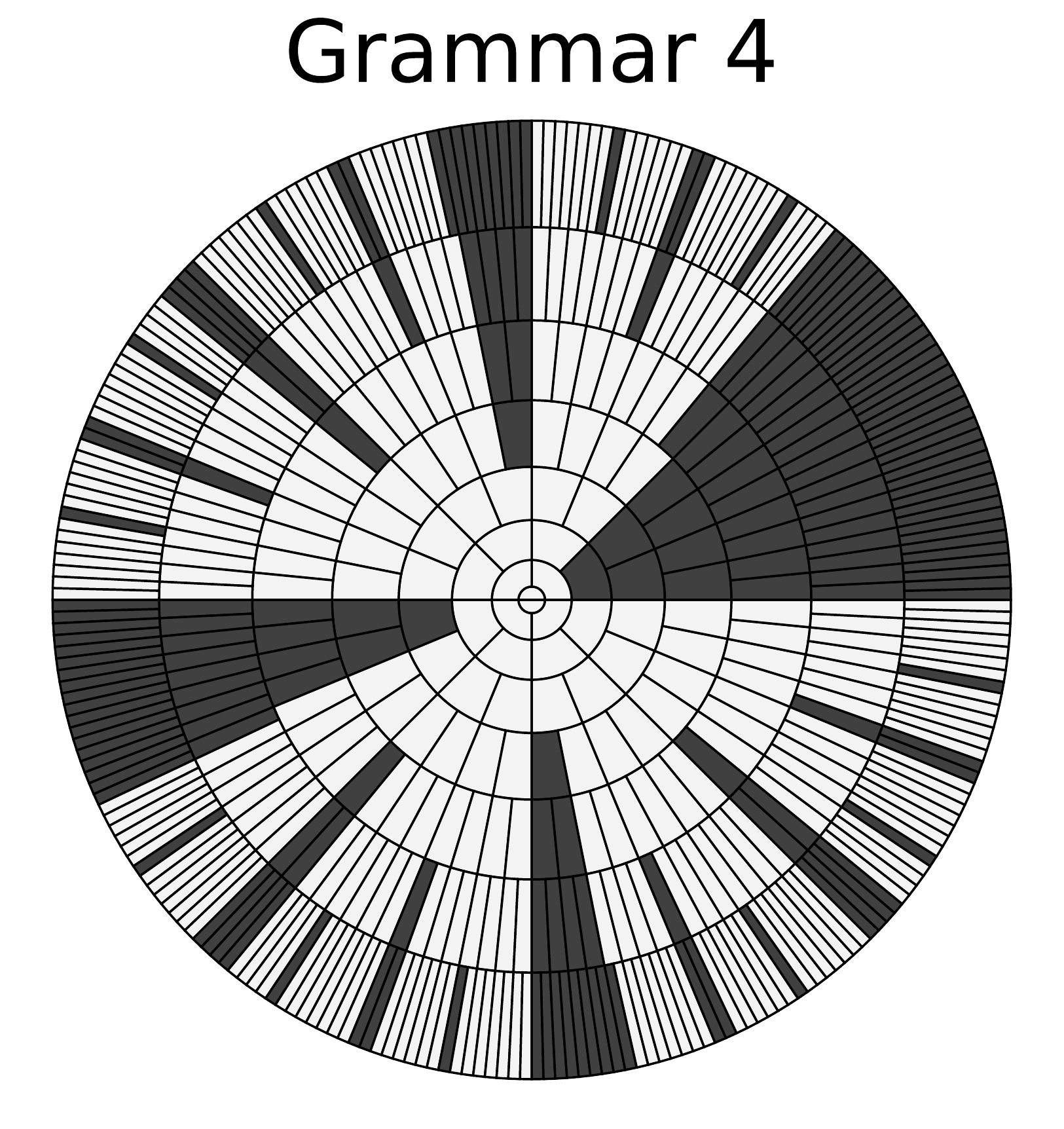}
  \vspace{-0.5em}
  \label{fig:g4_pie}
\end{subfigure} \hfill
\begin{subfigure}{.15\textwidth}
  \centering
  \includegraphics[width=\linewidth]{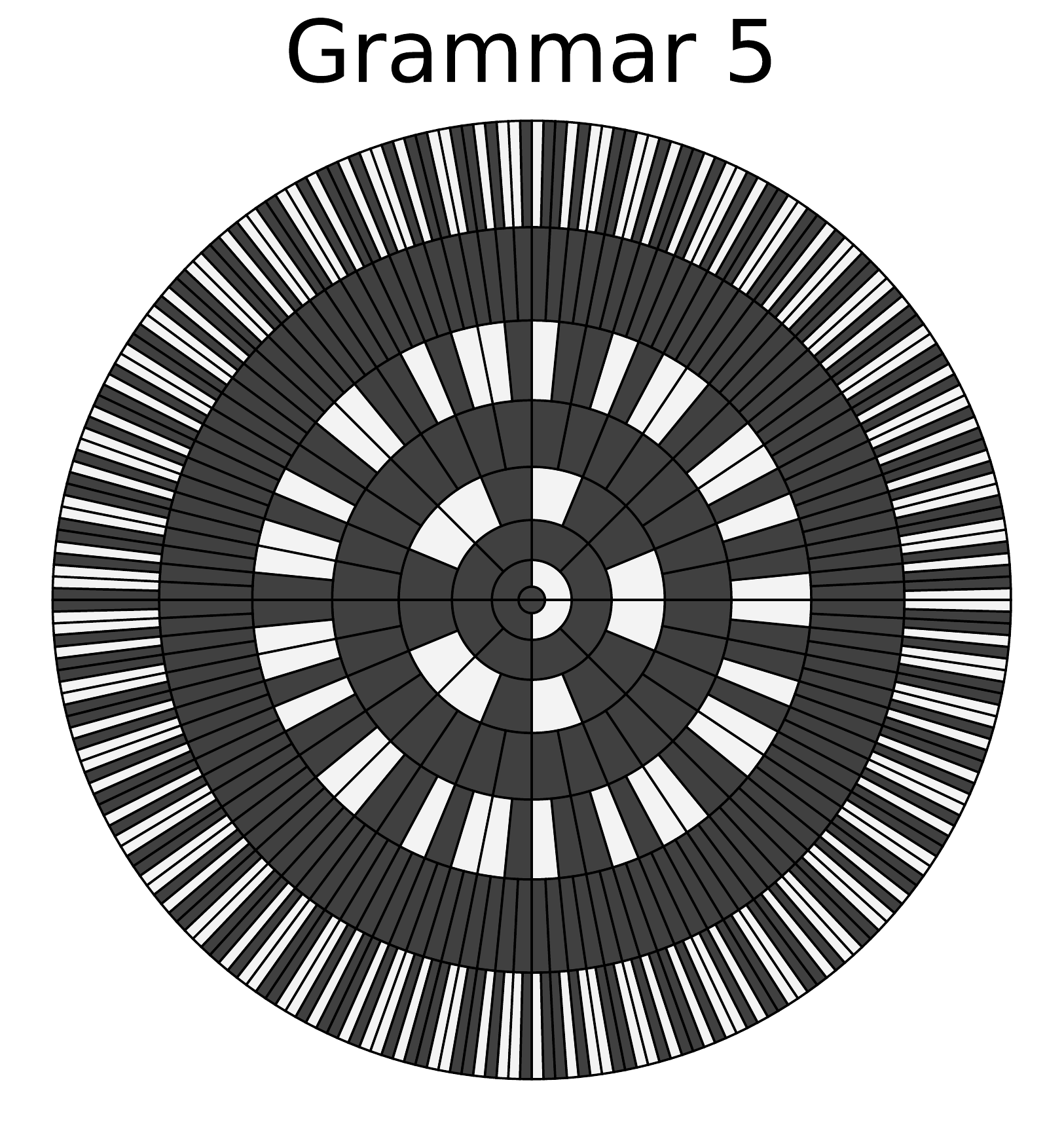}
  \vspace{-0.5em}
  \label{fig:g5_pie}
\end{subfigure} \hfill
\caption{Graphic presentation of the distribution of strings of length $N$ ($1 \leq N \leq 8$) for grammars 2, 4 and 5. In each concentric ring of either graph, there are $2^N$ strings arranged in lexicographic order, starting at $\theta = 0$. White and black areas represent positive and negative strings respectively.}
\label{fig:pie_grammars}
\end{figure}

Following prior work~\cite{watrous1992induction}, we plot two example graphs for grammar 2 and 5 in Figure~\ref{fig:pie_grammars} to better illustrate the differences between Tomita grammars.\footnote{The plots for other grammars are provided in the appendix.}. For each grammar, every concentric ring of its plot reflects the distribution of its associated positive and negative strings with a certain length (ranging from 1 to 8). Specifically, in each concentric ring, we arrange all strings with the same length in lexicographic order.\footnote{We do not impose any constraint on the order of string arrangement; other orders, e.g. gray codes, can also be selected for visualization.} As previously discussed in Section~\ref{sec:tomita}, the percentages of positive (or negative) strings for different grammars are very different. Especially, on grammar 5, the number of positive strings is equal to that of negative strings when the length of strings is even. Empirically, a data set consisting of balanced positive and negative samples should be desirable for training a model. However, as will be shown in the evaluation part of this study, this may make the learning difficult. For instance, when there are equal numbers of positive and negative strings for grammar 5, by flipping any binary digit of a string to its opposite (e.g. flipping a 0 to 1 or vice versa), any positive or negative string can be converted into a string with the opposite label. This implies that, in order to correctly recognizes all positive strings for grammar 5, a RNN will need to handle such subtle changes. Moreover, since this change can happen to any digit, a RNN must account for all digits without neglecting any. 

In the following, we introduce our defined entropy, followed by the definition of average edit distance. Then we show that these two metrics are closely related.

%% file: base/4_1_entropy.tex
\subsection{Entropy of Tomita Grammars}
\label{sec:entropy}
Given an alphabet $\Sigma = \{0 ,1\}$, we denote the collection of all $2^N$ strings of symbols from $\Sigma$ with length $N$ as $X^{N}$. For grammar $G$, let $X_{G_p}^{N}$ and $X_{G_n}^{N}$ represent the sets of positive and negative strings defined by $G$ in $X^{N}$, respectively. Then we have $X^{N} = X_{G_p}^{N} \cup X_{G_n}^{N}$. Let $m_{p}$ and $m_{n}$ denote the size of $X_{G_p}^{N}$ and $X_{G_n}^{N}$, i.e., $m_{p} = |X_{G_p}^{N}|$ and $m_{n} = |X_{G_n}^{N}|$, hence we have $m_{p} + m_{n} = 2^N$. The percentage of positive strings in $X^{N}$ is $r_{p} = m_{p} / 2^N$. To simplify the notation, here we use $X_{p}^{N}$ and $X_{n}^{N}$ to represent $X_{G_p}^{N}$ and $X_{G_n}^{N}$ respectively.

Assuming that all strings in $X^{N}$ are randomly distributed, we then denote the expected times of occurrence for an event $F$ -- two consecutive strings having different labels -- by $\mathrm{E} [F]$. We have the following definition of entropy for regular grammars with a binary alphabet.
\begin{definition}[Entropy]
\label{def:entropy}
Given a regular grammar $G$ with alphabet $\Sigma = \{0 ,1\}$, its entropy is defined as:
\begin{equation}
  \begin{aligned}
  H(G) = \underset{N \rightarrow \infty }{\mathrm{lim\,sup}} \,H^{N}(G)= \underset{N \rightarrow \infty }{\mathrm{lim\,sup}} \,\frac{1}{N}\, \log_{2}\, \mathrm{E} [F].      
  \end{aligned}
  \label{eq:entropy}
\end{equation}
where $H^N(G)$ is the entropy calculated for strings with the length of $N$.
\end{definition}

From Definition~\ref{def:entropy}, we have the following proposition: 
\begin{prop}
\label{prop:entropy}
	\begin{equation}
H(G)= 1 + \underset{N \rightarrow \infty }{\mathrm{lim\,sup}} \,\frac{\log_{2} \big(r_p(1 - r_p)\big)}{N}.
	\end{equation}
\end{prop}

A detailed proof is provided in the Appendix~\ref{sec:prop1_proof}. It is easy to find that following our definition, the value of entropy lies between $0$ and $1$. Based on the values of entropy for different grammars, we have the following theorem for categorizing regular grammars with a binary alphabet. 

\begin{thm} Given any regular grammar $G$ with alphabet $\Sigma = \{0 ,1\}$, it can be categorized into one of following classes:
\begin{enumerate}[label=(\alph*), wide, labelwidth=!, labelindent=0pt]

\item \emph{Polynomial} class. Any grammar $G$ in this class has the entropy $H(G) \!=\! 0$, if and only if the number of positive strings defined by $G$ has a polynomial form of $N$, i.e. $m_p \sim \mathrm{poly}(N)$.

\item \emph{Exponential} class. Any grammar $G$ in this class has the entropy $H(G) = \log_{2} b \in (0,1)$, if and only if the number of positive strings defined by $G$ has an exponential form of $N$, with the bases less than $2$, i.e. $m_p \sim \beta \cdot b^N$ where $b < 2$ and $\beta > 0$;

\item \emph{Proportional} class. Any grammar $G$ in this class has the entropy $H(G) \!=\! 1$, if and only if the number of positive strings defined by $G$ is proportional to $2^N$, i.e. $m_p \sim \alpha \cdot 2^N$, where $\alpha \in [0,1)$.
\end{enumerate}
\label{thm:entropy}
where $\sim$ indicates that some negligible terms are omitted when $N$ approaches infinity.
\end{thm}

For Tomita grammars, we categorize grammar 1, 2 and 7 into the polynomial class, grammar 3, 4 into the exponential class and grammar 5, 6 into the proportional class according the values of their entropy. A detailed proof for Theorem~\ref{thm:entropy} and the calculation of entropy for Tomita grammars are provided in Appendix~\ref{sec:appendix_entropy}. 

It should be noted that the concept of entropy has previously been introduced in the field of grammatical inference~\cite{carrasco1997accurate, thollard2000probabilistic}. The definition of entropy introduced in these studies is derived from information theory and is used to measure the relative entropy between stochastic regular grammars. An alternative definition of entropy that is closely related to our definition is introduced for measuring the ``information capacity'' of a wide class of shift spaces in symbolic dynamics~\cite{lind1995introduction}. More formally, the definition of shift space is as follows.
\begin{definition}[Shift Space]
\label{def:shift}
Given a full shift $A^{\mathbb{Z}}$, which is the collection of all bi-infinite sequences of symbols $x_i$ from $A$, denote a sequence as $x = (x_i)_{i \in \mathbb{Z}}$. A shift space $X$ is a subset of $A^{\mathbb{Z}}$ and $X = X_{\mathcal{F}}$ for some collection $\mathcal{F}$ of blocks that are forbidden over $A$. 
\end{definition}

An example shift space is the set of binary strings with no three consecutive 1's, i.e. $X = X_{\mathcal{F}}$, where $\mathcal{F} = \{111\}$. This shift space describes the same set of strings accepted by grammar 4. The entropy of a shift space is as follows.
\begin{definition}[Entropy of Shift Space]
\label{def:entropy_shift}
The entropy of a shift space $X$ is defined by:
\begin{equation}
  \begin{aligned}
  \label{eq:entropy_shift_space}
  H(X) = &\underset{N \rightarrow \infty }{\mathrm{lim}} \frac{1}{N}\, \log \, | \mathcal{B}_{N} (X) |
  \end{aligned}
  \vspace{-0.5em}
\end{equation}
where $| \mathcal{B}_{N} (X) |$ denotes the number of N-blocks in $X$.
\end{definition}
In Definition~\ref{def:entropy_shift} when the blocks and forbidden blocks of a shift space $X$ are regarded as positive and negative samples for $X$, $X$ can then be viewed as representing the data space described by a regular grammar. Despite that both Definition~\ref{def:entropy} and Definition~\ref{def:entropy_shift} can describe the complexity of $X$, Definition~\ref{def:entropy} is constructed by considering the distributions of both positive and negative strings, while Definition~\ref{def:entropy_shift} only considers positive strings. This more informative nature of Definition~\ref{def:entropy} has various benefits. For instance, when training a RNN and assuming a training set could coarsely reflect the real distributions of positive and negative samples, then by calculating the entropy according to Definition~\ref{def:entropy}, we can estimate the complexity of the entire data set. In addition, it is important to note Definition~\ref{def:entropy} can be easily generalized to a multi-class classification case, as one can always calculate the expected number of flips $\mathrm{E}[F] $ from samples. More formally, for a \emph{k}-class classification task with strings of length $N$, let $m_{i}$ denote the number of strings in the $i$th class. Then we have:
\begin{equation}
  \begin{aligned}
  \mathrm{E} [F] = 2^N -\frac{1}{2^N}\cdot \sum_{i=1}^{k}  m_{i}^{2}.
  \end{aligned}
\label{eq:E_runs_pn}
\end{equation}
By substituting~\eqref{eq:E_runs_pn} in Definition~\ref{def:entropy}, we are able to compute the entropy for this \emph{k}-class classification task. As for Definition~\ref{def:entropy_shift}, it only defines the entropy in an one-versus-all manner. Moreover, it should be noted that not all regular grammars can have their data space be represented as a shift space, especially for grammars that lack the shift-invariant and closure properties~\cite{lind1995introduction}. As such, Definition~\ref{def:entropy_shift} cannot be applied for estimating the complexity of these grammars.

%% file: base/4_2_distance.tex
\subsection{Average Edit Distance of Tomita Grammars}
\label{sec:avg_edit_dist}
We now formally define the average edit distance of regular grammars with a binary alphabet in order to measure the difference between the sets of positive and negative strings for a given grammar $G$. We first revisit the definition of \emph{edit} distance~\cite{de2010grammatical}, which measures the minimum number of operations -- substituting one symbol for another, or flipping a ``1'' to ``0'' or vice versa in our case\footnote{The operations include insertion, deletion and substitution of one symbol from a string. It should be noticed that since we fix the length of strings, we omit the operations including insertion and deletion and only consider substitution.} -- needed to covert a positive or negative string into another.

\begin{definition}[Edit Distance]
\label{def:distance}
Given two strings $x$ and $y$ in $X^N$, $x$ rewrites into $y$ in a one-step operation if the following single-symbol substitution holds:
\begin{equation}
  \begin{aligned}
  \label{eq:edit_dist}
  \vspace{-0.5em}
  x = uav, \; y = ubv \; \text{and} \; u, v \in X^N, \; a, b \in A. \nonumber
  \end{aligned}
\end{equation}
Let $x \overset{k}{\rightarrow} y$ denote $x$ rewrites into $y$ by $k$ operations of a single-symbol substitution. Then the edit distance between $x$ and $y$ denoted by $d_{edit}(x, y)$ is the smallest $k$ such that $x \overset{k}{\rightarrow} y$.
\end{definition}

Since we only consider single-symbol substitution, in our case $k$ is equal to the \emph{Hamming} distance between $x$ and $y$. In the following, we expand Definition~\ref{def:distance} to calculate the average edit distance between the set of positive strings, $X_{p}^N$ and the set of negative strings, $X_{n}^N$ for grammar $G$. Given a positive string $x_p \in X_{p}^N$ and a negative string $x_n \in X_{n}^N$, the edit distance between $x_p$ and all negative strings, and the edit distance between $x_n$ and all positive strings can be expressed as: 
\begin{equation}
  \begin{aligned}
  \label{eq:dist_neg_pos}
  d_{edit}(x_p, X_{n}^N) &= \underset{x_n \in X_{n}^N}{\mathrm{min}} d_{edit}(x_p, x_n), \\
  d_{edit}(x_n, X_{p}^N) &= \underset{x_p \in X_{p}^N}{\mathrm{min}} d_{edit}(x_n, x_p).
  \end{aligned}
\end{equation}
Then we have the following definition of average edit distance:
\begin{definition}[Average Edit Distance]
\label{def:distance_avg}
Given a grammar $G$, the average edit distance $D(G)$ between the positive and negative strings defined by $G$ is: 
\begin{equation}
  \begin{aligned}
  \label{eq:distance_g}
  D(G) = \frac{1}{2} \cdot \underset{N \rightarrow \infty }{\mathrm{lim}}  D^N(G),
  \end{aligned}
\end{equation}
where
\begin{equation}
  \begin{aligned}
  \label{eq:distance_g_n1}
  D^N(G) = \frac{1}{|X_{p}^N|} D_{p}^{N} + \frac{1}{|X_{n}^N|} D_{n}^{N}
  \end{aligned}
\end{equation}
calculates the average edit distance for strings with their length equal to $N$. $D_{p}^{N}$ and $D_{n}^{N}$ denote $\sum_{x_P \in X_{p}^N} d_{edit}(x_p, X_{n}^N)$ and $\sum_{x_n \in X_{n}^N} d_{edit}(x_n, X_{p}^N)$, respectively.
\end{definition}
Using Definition~\ref{def:distance_avg}, we can categorize Tomita grammars into three classes that are identical to the classes previously introduced in Theorem~\ref{thm:entropy}. Detailed calculation of the average edit distance for each grammar is provided in the Appendix~\ref{sec:appendix_dist}.

\begin{enumerate}[label=(\alph*), wide, labelwidth=!, labelindent=0pt]
  \item For grammar 1, 2 and 7, $D(G_{1,2,7}) = \infty$; 
  \item For grammar 3 and 4, $D(G_{3,4}) > 1$;
  \item For grammar 5 and 6, $D(G_{5,6}) = 1$.
\end{enumerate}

%% file: base/4_3_entropy_dist.tex
\subsection{Relationship Between Entropy and Average Edit Distance}
\label{sec:entropy_dist}
By comparing the values of entropy and average edit distance defined for Tomita grammars, it is evident that these two defined metrics are closely related to each other. In particular, while the entropy of grammar 5 and 6 has the maximum value of 1, their average edit distance has the minimum value of 1. Additionally, the entropy of grammar 1, 2 and 7 has the minimum value of 0, while their average edit distance approaches infinity as $N$ increases. 

More formally recall that in~\eqref{eq:distance_g_n1}, $D_{p}^{N}$ and $D_{n}^{N}$ calculate the summed edit distance for all positive strings and all negative strings, respectively, when the length of strings is $N$. Then we have the following proposition.
\begin{prop}
\label{prop:relation}
	\begin{equation}
    H(G) = \underset{N \rightarrow \infty }{\mathrm{lim\,sup}} \frac{1}{N} \cdot \log_2 (r_p D_{n}^{N} + (1-r_p) D_{p}^{N}) .
	\end{equation}
\end{prop}

A detailed proof is provided in Appendix~\ref{sec:appendix_prop_2}. Assuming that a random variable $\tilde{D}_{v}^{N}$ takes a value from $\{D_{n}^{N}, D_{p}^{N}\}$ with the probability of $r_{p}$ for selecting $D_{n}^{N}$ and $1 - r_{p}$ for selecting $D_{p}^{N}$, then Proposition~\ref{prop:relation} calculates $\mathrm{E}[\tilde{D}_{v}^{N}]$, which is the expected summation of the edit distance between positive and negative strings for any grammar $G$. 

\begin{table}[t]
\centering
\small
\caption{Entropy \& avg. edit distance for Tomita grammars.}
\label{tab:computed_H_D}
\begin{tabular}{ccccccccc}
\hline\hline
\multirow{2}{*}{Def.} & \multirow{2}{*}{Len.} & \multicolumn{7}{c}{Grammar}                    \\ \cline{3-9} 
                      &                       & G1   & G2   & G3   & G4   & G5   & G6   & G7   \\ \hline\hline
\multirow{4}{*}{$H^{N}$} & 8                   & 0.12 & 0.12 & 0.87 & 0.87 & 0.88 & 0.85 & 0.86 \\
                      & 10                  & 0.10 & 0.10 & 0.88 & 0.89 & 0.90 & 0.88 & 0.82 \\
                      & 12                  & 0.08 & 0.08 & 0.88 & 0.91 & 0.92 & 0.90 & 0.76 \\
                      & 14                  & 0.07 & 0.07 & 0.88 & 0.92 & 0.93 & 0.92 & 0.70 \\ \hline
\multirow{4}{*}{$D^{N}$} & 8                   & 2.51 & 2.51 & 1.13 & 1.16 & 1.00 & 1.00 & 1.17 \\
                      & 10                  & 3.00 & 3.00 & 1.18 & 1.16 & 1.00 & 1.00 & 1.31 \\
                      & 12                  & 3.50 & 3.50 & 1.24 & 1.18 & 1.00 & 1.00 & 1.51 \\
                      & 14                  & 4.00 & 4.00 & 1.30 & 1.22 & 1.00 & 1.00 & 1.75 \\ \hline\hline
\end{tabular}
\end{table}

In Table~\ref{tab:computed_H_D}, we show the values of entropy and average edit distance for each grammar calculated by varying the length $N$ of generated strings from 8 to 14. Clearly, as $N$ increases, the entropy of grammars 1, 2 and 7 monotonically decreases while their average edit distance monotonically increases. For other grammars, both their entropy and average edit distance change in the directions that are opposite to that for grammar 1, 2 and 7. In addition, the results shown in Table~\ref{tab:computed_H_D} also demonstrate the difference between entropy and average edit distance. Specifically, when only observing the entropy, it is difficult to distinguish grammars 3 and 4 from grammars 5 and 6. The difference between these two classes of grammars is more clearly demonstrated when comparing their average edit distance. In particular, the average edit distance of grammars 5 and 6 is constantly equal to 1, while the average edit distance for grammars 3 and 4 keeps increasing as $N$ increases. This indicates that average edit distance reveals more information about a regular grammar when compared with entropy. However, it is also important to note that the time and space cost for calculating average edit distance can be significantly higher than that needed for calculating entropy. Thus there is a trade-off between the granularity and computational efficiency when using these two metrics.

%% file: base/5_0_exp_config.tex
\section{Experiments}
\label{sec:eval}
In this section, we first present our experiment setup, including the data sets generated and configurations of the recurrent networks selected. We then introduce the DFA extraction procedure adopted in this study. Last, we provide experimental results and discussion. 

\subsection{Data Set}
\label{sec:data}

We followed the approach introduced by~\cite{giles1992learning, Empirical2017Wang} to generate string sets for Tomita grammars. To be specific, we drew strings from the grammar specified in Table~\ref{tab:tomita} and an oracle generating random 0 and 1 strings. The end of each string is set to symbol 2, which represents the ``stop'' symbol (or end-token as in language modeling). The strings drawn from a grammar were designated as positive samples while those from that random oracle as negative samples. Note that we verified each string from the random oracle to ensure they are not in the string set represented by that corresponding grammar before treating them as negative samples. It should be noticed that each grammar in our experiments represents one set of strings with an unbounded size. As such we restricted the length of the strings drawn within a certain range (listed in the column ``Length'' of Table~\ref{tab:tomita}). In our experiments, we specify a lower bound on the string lengths to avoid training RNNs with empty strings. In order to use as many strings as possible to build the datasets, the lower bound should be set to be sufficiently small. We set the lower bound equal to the minimal number of states presented in the corresponding DFA, and the upper bound to allow each state of a DFA to be at least visited twice. In particular, for grammar 1, 2 and 7, it can be easily checked that the data sets for these three grammars to be very imbalanced for positive and negative samples. We up-sampled positive strings in our experiments for these three grammars for the training of RNNs.

We split the strings generated within the specified range of length for each grammar to build the training and testing sets according to the ratios listed in Table~\ref{tab:tomita}. Both training and testing sets were used to train and test the RNNs accordingly, while only the testing sets were used for evaluating the extracted DFAs.

\subsection{Recurrent Networks Setup}
\label{sec:model_setup}

\begin{table*}[t]
\centering
\caption{Description of recurrent networks investigated.}
\label{tab:rnn_models}
\begin{tabular}{lll}
\hline\hline
Model                                                      & Hidden Layer Update                                                                                                                                                                                                                           & Parameters                                                         \\ \hline\hline
Elman-RNN~\cite{elman1990finding}                                                  & $s^{t+1} = \phi(U I^t + W s^t + b)$                                                                                                                                                                                                               & $U \in \mathbb{R}^{N_s \times N_I}, W \in \mathbb{R}^{N_s \times N_s}$               \\\hline
\begin{tabular}[c]{@{}l@{}}Second-order\\ RNN~\cite{giles1992learning}\end{tabular} & $s_{i}^{t+1} = \phi(\sum_{j,k} W_{ijk} s_{j}^{t} I_{k}^{t}), \; i,j = 1 \dots N_s, \; k = 1 \dots N_I$                                                                                                                                                                                                         & $W \in \mathbb{R}^{N_s \times N_s \times N_I}$                                 \\ \hline
MI-RNN~\cite{wu2016multiplicative}                                                     & $s^{t+1} = {\tt tanh} (\alpha \otimes U I^{t} \otimes W s^{t} + \beta_{1} \otimes W s^{t} + \beta_{2} \otimes U I^t + b)$                                                                                                                                                      & \begin{tabular}[c]{@{}l@{}} $U \in \mathbb{R}^{N_s \times N_I}, W \in \mathbb{R}^{N_s \times N_s}$, \\ $\alpha \in \mathbb{R}^{N_s}, \beta_1 \in \mathbb{R}^{N_s}, \beta_2 \in \mathbb{R}^{N_s}$ \end{tabular} \\ \hline
LSTM~\cite{hochreiter1997long}                                                       & \begin{tabular}[c]{@{}l@{}}$i^{t+1} = \sigma(U_i I^t + W_i s^{t})$, $f^{t+1} = \sigma(U_f I^t + W_f s^{t})$, \\ $o^{t+1} = \sigma(U_o I^t + W_o s^{t})$, $g^{t+1} = {\tt tanh}(U_g I^t + W_g s^{t})$ \\ $c^{t+1} = c^{t} \otimes f^{t+1} + g^{t+1} \otimes i^{t+1}$, $s^{t+1} = {\tt tanh}(c_{t+1}) \otimes o^{t+1}$ \end{tabular} & \begin{tabular}[c]{@{}l@{}} $U_{*} \in \mathbb{R}^{N_s \times N_I}, W_{*} \in \mathbb{R}^{N_s \times N_s}$ \\ $* = \{i, f, o, g \}$\end{tabular}                                                              \\ \hline
GRU~\cite{cho2014properties}                                                        & \begin{tabular}[c]{@{}l@{}}$z^{t+1} = \sigma(U_z I^t + W_z s^{t})$,  $r^{t+1} = \sigma(U_r I^t + W_r s^{t})$,\\ $h^{t+1} = {\tt tanh} (U_h I^t + W_h (s^{t} \otimes r^{t+1}))$, $s^{t+1} = (1-z^{t+1}) \otimes h^{t+1} + z^{t+1} \otimes s^{t}$ \end{tabular}                                               & \begin{tabular}[c]{@{}l@{}} $U_{*} \in \mathbb{R}^{N_s \times N_I}, W_{*} \in \mathbb{R}^{N_s \times N_s}$ \\ $* = \{z, r, h \}$\end{tabular}                                                              \\ \hline\hline
\end{tabular}

\end{table*}

Here we provide an unified view of the update activity of recurrent neurons for the recurrent networks used here. We investigated Elman-RNN, second-order RNN, MI-RNN, LSTM and GRU RNNs. These models were selected based on whether they were frequently adopted either in previous work on DFA extraction or in recent work on processing sequential data.

A recurrent model consists of a hidden layer $s$ containing $N_s$ recurrent neurons (an individual neuron designated as $s_i$), and a input layer $I$ containing $N_I$ input neurons (each designated as $I_k$). We denote the values of its hidden layer neuron at $t\,$th and $t+1\,$th discrete times as $s^{t}$ and $s^{t+1}$. Then the hidden layer is updated by:
\begin{equation}
  \begin{aligned}
  \label{eq:basic_rnn}
  s^{t+1} = \phi(I^t, s^{t}, W_P), \nonumber
  \end{aligned}
\vspace{-0.5em}
\end{equation}
where $\phi$ is the nonlinear activation function, and $W_P$ denotes the weight parameters which modifies the strength of interaction among input neurons, hidden neurons, output neurons and any other auxiliary units. In most recurrent networks, the weight parameters $W_P$ usually comprise two separate weights, i.e. $U \in \mathbb{R}^{N_s \times L}$ and $W \in \mathbb{R}^{N_s \times N_s}$. Then inputs $I^t$ and hidden value $s^{t}$ at the $t\,$th discrete time are multiplied by weight $U$ and $W$, respectively. Due to space constraints, a detailed description of the hidden layer update for each model is presented in Table~\ref{tab:rnn_models}.

\paragraph{Activation Function} 
Here we follow previous research~\cite{giles1992learning}, which mainly uses either activation functions -- {\tt sigmoid} and {\tt tanh} -- to build recurrent networks. We choose both for Elman and second-order RNNs. We do not evaluate the impact of {\tt ReLU} upon DFA extraction even thought it has been broadly applied for recurrent networks. This is due to the fact that DFA extraction needs to perform hidden vector clustering and the range of the {\tt ReLU} function between 0 and infinity makes hidden vector clustering not obvious.

\paragraph{Model Parameters} 
We used recurrent networks with approximately the same number of weight and bias parameters (shown in Table~\ref{tab:tomita}). Specifically, with Elman-RNN and second-order RNN as examples, we denote the number of hidden neurons for these models as $N_{s_1}$ and $N_{s_2}$, and denote the number of weight parameters of these models as $N_{W_1}$ and $N_{W_2}$, respectively. Then we have $N_{W_1} = N_{s_1}\times N_{s_1} + N_{s_1}\times N_{I} + N_{s_1}$ and $N_{W_2} = N_{s_2}\times N_{s_2}\times N_{I}$. By setting $N_{W_1} \!\! \approx \!\!N_{W_2}$, we then determine the size of the hidden layer for each model.

\paragraph{Input Encoding} 
For each model, we use one-hot encoding to process the input symbols. With this configuration, the input layer is designed to contain a single input neuron for each character in the alphabet of the target language. Thus, only one input neuron is activated at each time step.

\paragraph{Loss Function}
We follow the approach introduced in~\cite{giles1992learning} and apply the following loss function to all recurrent networks:
\vspace{-0.3em}
\begin{equation}
  \begin{aligned}
  Loss = \frac{1}{2}(y - s_{T}^{0})^{2}. \nonumber
  \end{aligned}
\vspace{-0.3em}
\end{equation}
This loss function is viewed as selecting a special ``response'' neuron $s^{0}$ and comparing it to the label $y$. $s_{T}^{0}$ indicates the value of $s^0$ at time $T$ after a model receives the final input symbol. By using this simple loss function, we expect to eliminate the potential effect of adding an extra output layer and introducing more weight and bias parameters. Through this design, we can ensure the knowledge learned by a model resides in the hidden layer and its transitions. During training, we optimize parameters through stochastic gradient descent and employ the \textit{RMSprop} adaptive learning rate scheme~\cite{tieleman2012lecture}.

\subsection{Procedure of DFA Extraction}
\label{sec:extraction}

\paragraph{Configuration of DFA Extraction}
\label{sec:recipe_extraction}
Recall the basic procedure for DFA extraction introduced in Section~\ref{sec:background}. Here we specify our configurations for each step as follows:
\begin{enumerate}[wide, labelwidth=!, labelindent=0pt]
	\item By collecting all hidden vectors computed by a RNN on all strings from a data set generated as previously discussed, we quantize the continuous space of hidden vectors into a discrete space consisting of a finite set of states. In most previous research, this quantization is usually implemented with clustering methods including k-means clustering~\cite{zeng1993learning, frasconi1996representation, gori1998inductive}, hierarchical clustering~\cite{sanfeliu1994active}, and self-organizing maps~\cite{tivno1995learning}. Here, we use k-means due to its simplicity and computational efficiency.
    \item With each hidden vector assigned to a unique cluster, we construct a state transition table. In prior studies, this is conducted by breadth-first search (BFS)~\cite{giles1992learning} or sampling approaches~\cite{tivno1995learning}. A survey of these methods~\cite{jacobsson2005rule} shows that BFS approaches can construct a transition table relatively consistently but incurs high computational cost when the size of alphabet increases. Compared with BFS approaches, sampling approaches are more computationally efficient. However, they may introduce inconsistencies when constructing a transition table. To achieve a trade-off between computation efficiency and construction consistency, we follow~\cite{schellhammer1998knowledge,Empirical2017Wang} and count the number of transitions observed between states. Then we only preserve the more frequently observed transitions.
    \item With a transition diagram constructed, we utilize a standard and efficient DFA minimization algorithm~\cite{hopcroft2006automata} which has been broadly adopted in previous works for minimizing DFAs extracted from different recurrent networks and for other DFA minimization.
\end{enumerate}

\paragraph{Random Trials for DFA Extraction}
\label{sec:random_trials}
In order to more comprehensively evaluate the performance of DFA extraction for different recurrent networks trained on different grammars, in our experiments we perform multiple trials of DFA extraction for each RNN on every grammar. In particular, given a RNN and the data set generated by a grammar, we vary two factors -- the initial value of the hidden vector of this RNN and the pre-specified value of $K$ (indicating the number of clusters) for k-means clustering that performs the DFA extraction. In our prior study~\cite{Empirical2017Wang}, we empirically demonstrate that for a sufficiently well trained (100.0\% accuracy on the training set) second-order RNN, the initial value of hidden layer has significant influence on the extraction performance when $k$ is set to small values. This impact can be gradually alleviated when $K$ increases. We observer that when $k$ is sufficiently large, the influence of randomly initializing the hidden layer is negligible. As such, in our experiments for every pair of a recurrent model and a grammar, we conducted 10 trials with random initialization of the hidden layer of that model~\footnote{For each trial, we select a different seed for generating the initial hidden activations randomly.}. Within each trial, we train this recurrent model on the training set associated with this grammar until convergence. Then we apply DFA extraction on this model multiple times by ranging $K$ from 3 to 15. In total we perform DFA extraction 130 times for each model on each grammar. We tested and recorded the accuracy of each extracted DFA using the same test set constructed for evaluating the corresponding recurrent model. The extraction performance is then evaluated based on results obtained from these trials of extraction. Through this, we believe we alleviate the impact of different recurrent networks being sensitive to certain initial state settings and clustering configurations.

%% file: base/5_1_acc.tex
\subsection{Comparison of the Quality of the Extracted DFAs}
\label{sec:acc}
\begin{figure*}[t]
\centering
\begin{subfigure}{.28\textwidth}
  \centering
  \includegraphics[width=0.95\linewidth]{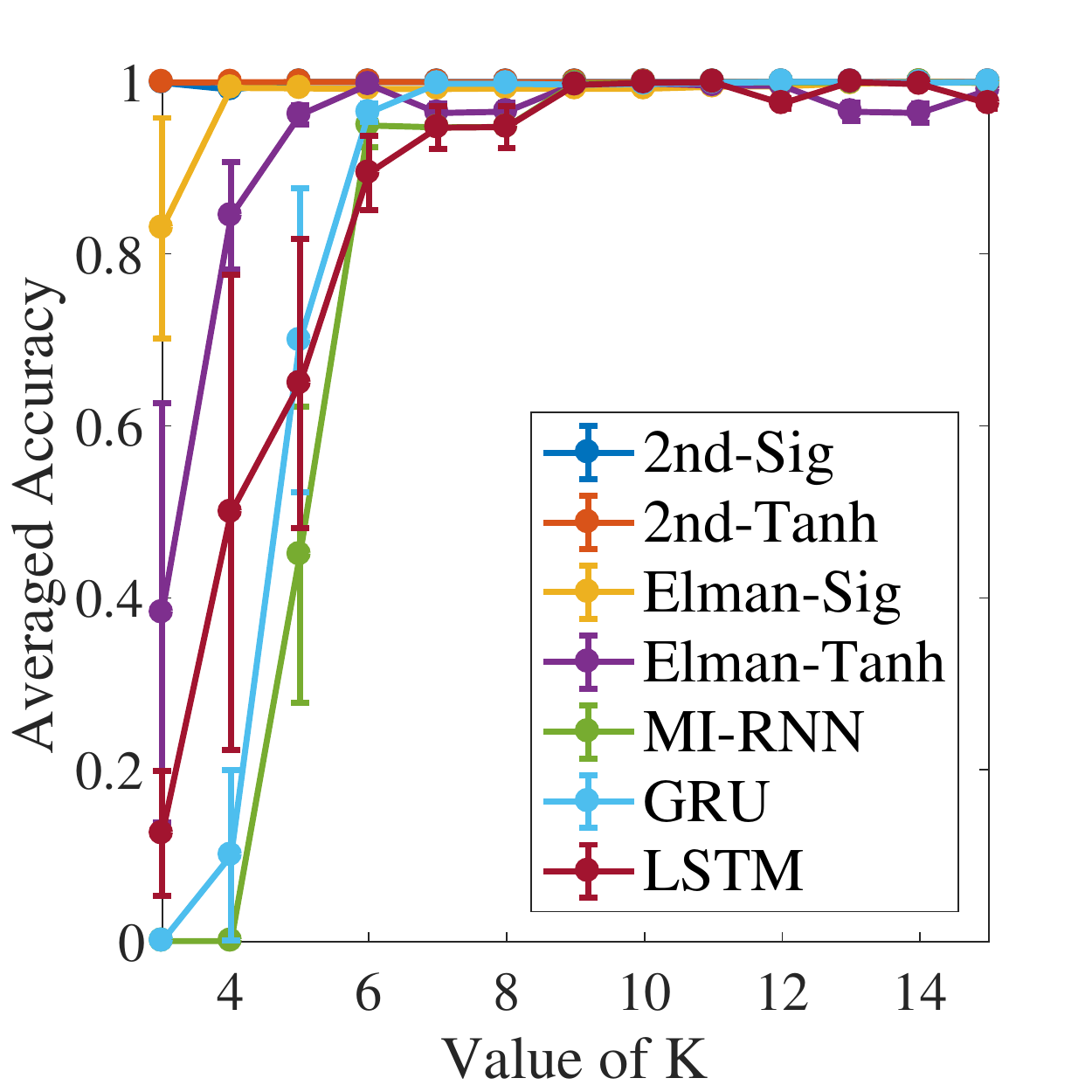}
  \centering
  \vspace{-0.5em}
  \caption{Grammar 2.}
  \label{fig:g2_line}
\end{subfigure} \hfill
\begin{subfigure}{.28\textwidth}
  \centering
  \includegraphics[width=0.95\linewidth]{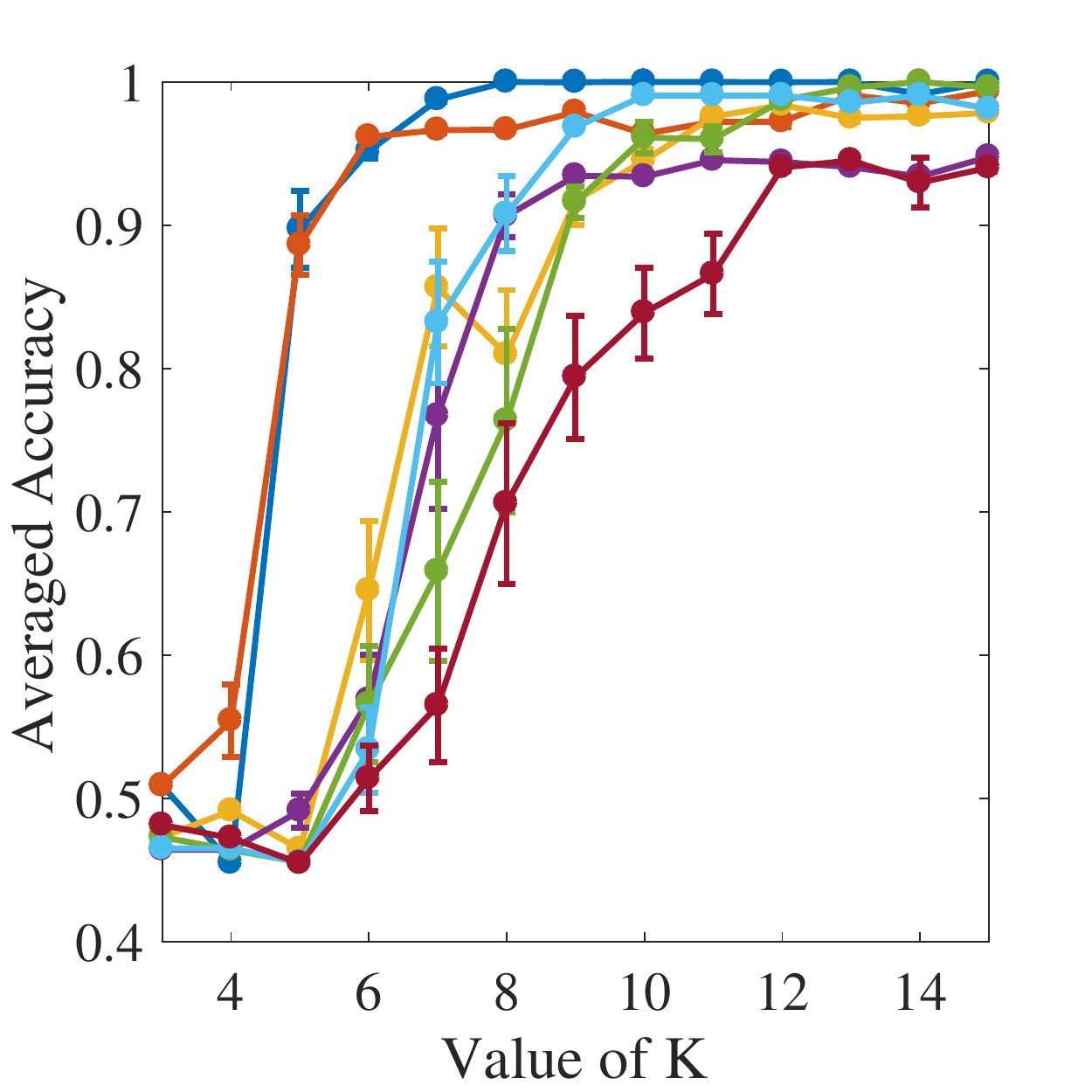}
  \vspace{-0.5em}
  \caption{Grammar 4.}
  \label{fig:g4_line}
\end{subfigure} \hfill 
\begin{subfigure}{.28\textwidth}
  \centering
  \includegraphics[width=0.95\linewidth]{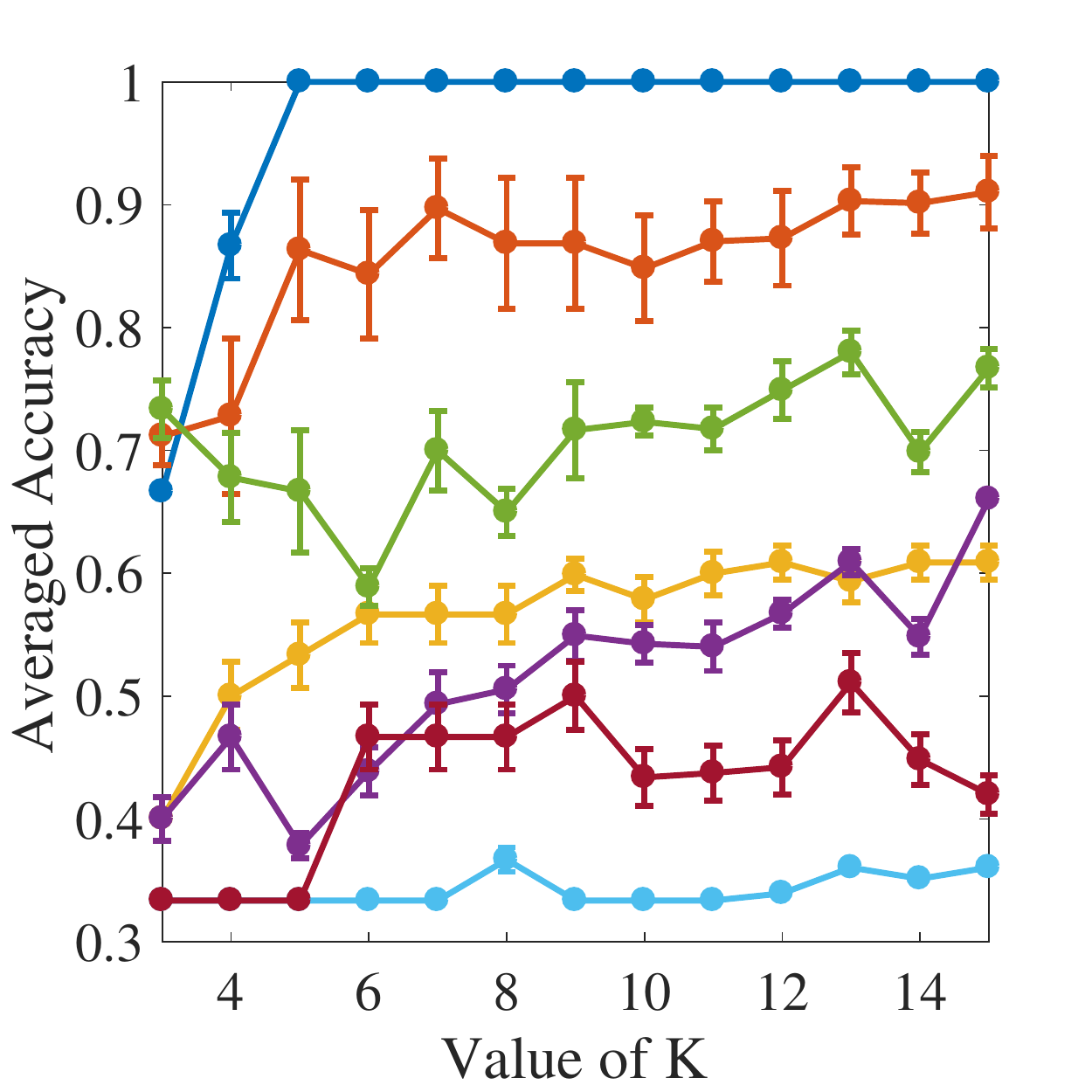}
  \vspace{-0.5em}
  \caption{Grammar 5.}
  \label{fig:g5_line}
\end{subfigure} \hfill
\caption{Mean and variance of the accuracy obtained by DFAs extracted from all models on grammar 2, 4 and 5. We denote second-order RNN with {\tt sigmoid} and {\tt tanh} activation function by 2nd-Sig and 2nd-Tanh. Similarly, Elman-RNN with these two activation functions are denoted as Elman-Sig and Elman-Tanh respectively.}
\label{fig:avg_var}
\end{figure*}
In the following experiments, we evaluate and compare the quality of DFAs extracted from different RNNs trained on different Tomita grammars. All models are trained to achieve 100.0\% accuracy on the training sets constructed for all grammars. Particularly, LSTM and GRU converge much faster than other models. This is as expected, since the data sets are rather simple in comparison with more sophisticated sequential data, e.g. natural language and programming code, on which LSTM and GRU have demonstrated impressive successes. The training results are omitted here due to space constraints. 

Given a particular recurrent model and a grammar, the quality of extracted DFAs is evaluated by calculating from multiple trails the mean and variance of the accuracy obtained by extracted DFAs. In Figure~\ref{fig:avg_var}, we show the results for grammar 2, 4 and 5 (as representatives for the three categories of Tomita grammars introduced in Section~\ref{sec:complexity}). For each category of grammars, the results for other grammars are similar to the results obtained from these three representative grammars and are provided in the Appendix. In Figure~\ref{fig:avg_var}, we observe that on grammars with lower complexity, i.e. grammar 2 and 4, different models behave similarly. Specifically, all models produce DFAs with gradually increasing accuracy and decreasing variance of accuracy, and eventually produce DFAs with near or equal to 100.0\% accuracy. It is clear that random initialization of the hidden layer has an impact on the quality of extracted DFAs only when $K$ is relatively small and is alleviated when $K$ is sufficiently large. In particular, it can be noticed that among all extracted DFAs, DFAs extracted from second-order RNN achieve the highest accuracy with the lowest variance. Upon closer examination, we observe that the values of $K$ needed by second-order RNNs for extracting DFAs with near or equal to 100.0\% accuracy are smaller than those needed by other models. 

The quality of DFAs extracted from different recurrent models is rather diverse for more complex grammars. More specifically, on grammar 5, only second-order RNNs with {\tt sigmoid} and {\tt tanh} activation are able to extract DFAs that achieve 100.0\% accuracy, while all other models fail. This reflects that all other models, except for second-order RNN, are sensitive to the complexity of the underlying grammars on which they are trained. In particular, DFAs extracted from Elman-RNN with {\tt sigmoid} and {\tt tanh} activation, LSTM, and GRU perform much worse that other models. For the Elman-RNN, its worse extraction performance may due to its simple recurrent architecture which somehow limits its ability to capture complicated symbolic knowledge. However, the worse results obtained by LSTM and GRU on grammar 5 are surprising. One possible explanation is that for a recurrent model with a more complicated update activity for its hidden layer, the vector space of its hidden layer may not be spatially separable. As a result, clustering methods developed based on Euclidean distance, including k-means, could not effectively identify different states. Instead, for LSTM and GRU, the gate units constructed in these models might function as decoders for recognizing states that are not spatially separated. Nevertheless, it is an open question on how to extract DFA from these more complicated models.
\begin{figure}[t!]
\hfill
\begin{center}
\vspace{-1.0em}
\includegraphics[width=0.95\linewidth]{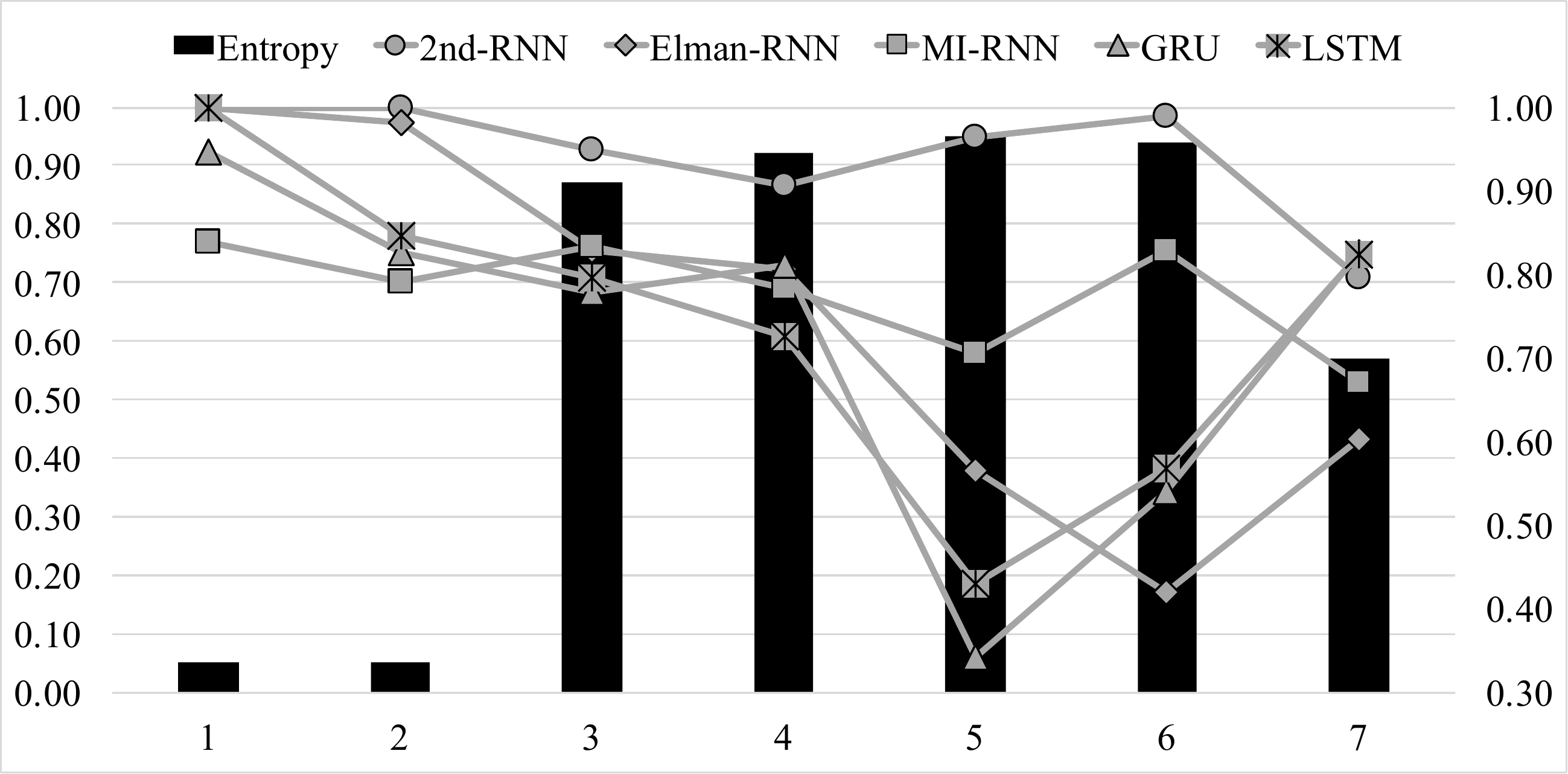}
\end{center}
\vspace{-1.0em}
\caption{Average accuracy of DFAs extracted from recurrent models on Tomita grammars. Left vertical axis: entropy. Right vertical axis: average accuracy of extracted DFAs.}
\label{fig:acc_entropy}
\end{figure}

To better illustrate the influence of different grammars on the quality of extracted DFAs, Figure~\ref{fig:acc_entropy} plots the average accuracy of 130 DFAs extracted from each model trained on each grammar, and the entropy of each grammar calculated by setting $N = 20$. We only show the the results for second-order RNN and Elman-RNN with a {\tt sigmoid} activation function since the results obtained by these models with the {\tt tanh} activation are usually worse. As shown in Figure~\ref{fig:acc_entropy}, except for second-order RNN and MI-RNN, the average accuracy obtained by DFAs extracted from each model decreases as the entropy of grammars increases. This result indicates that it is generally more difficult for recurrent models to learn a grammar with higher level of complexity. In general, DFAs extracted from second-order RNN have consistently higher accuracy across all grammars. This better performance of second-order RNN on DFA extraction raises questions regarding the quadratic interaction between input and hidden layers used by this model and whether such an interaction could improve other models DFA extraction.

%% file: base/5_2_success.tex
\subsection{Comparison of the Success Rate of DFA Extraction}
\label{sec:success}

\begin{table*}
\begin{minipage}[b]{0.75\linewidth}
\centering
  \includegraphics[width=0.85\linewidth]{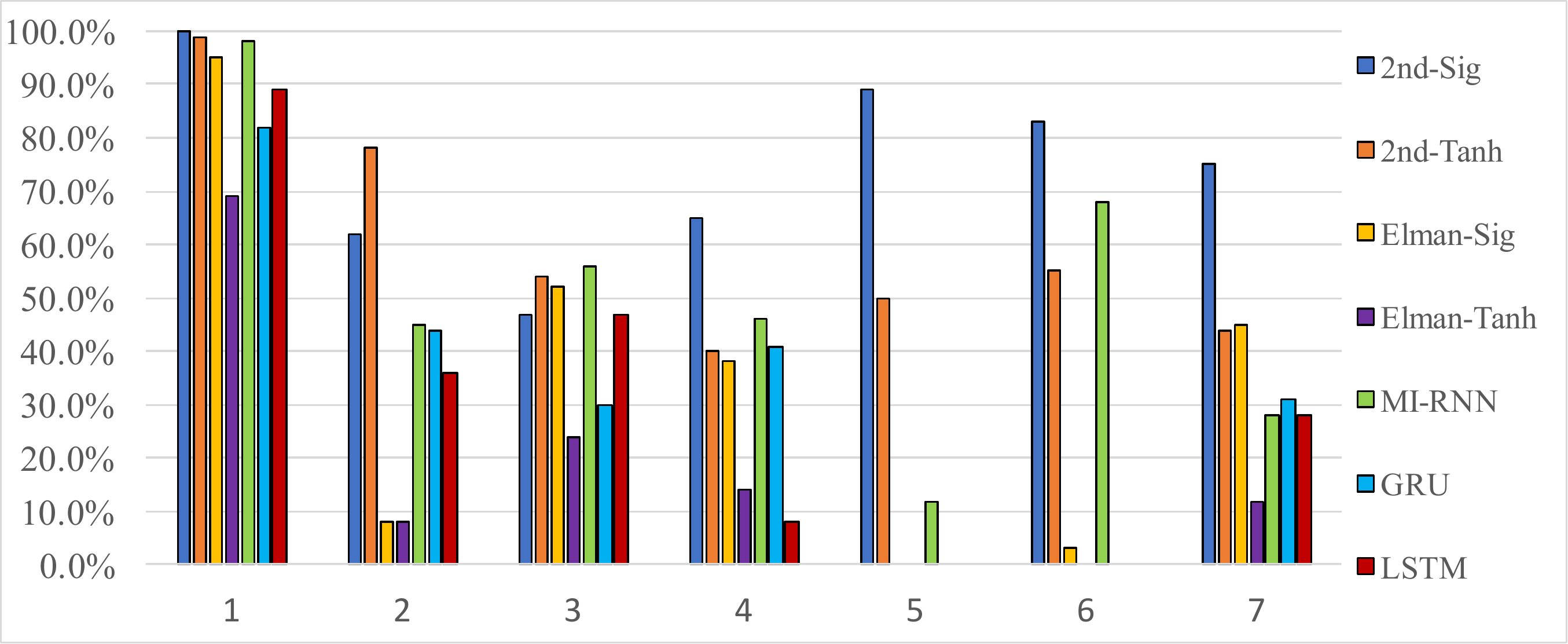}
  \captionof{figure}{Success rates of DFA extraction for all models on Tomita grammars.}
  \label{fig:g_bar_all}
\end{minipage}\hfill
\begin{minipage}[b]{0.20\linewidth}
\centering
  \includegraphics[width=0.7\linewidth]{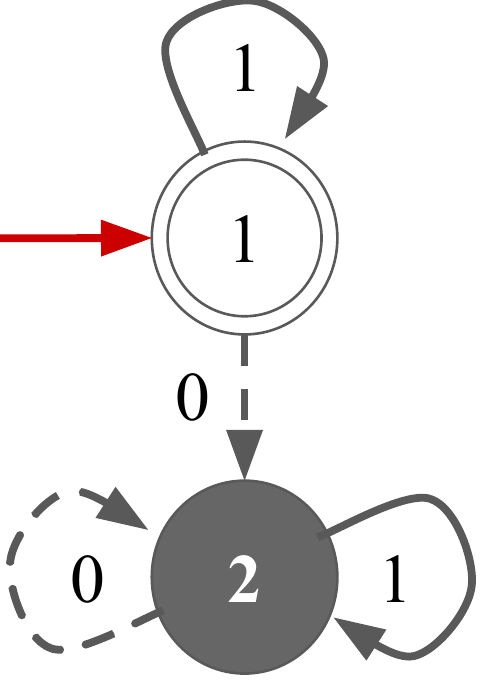}
  \captionof{figure}{DFA example for Tomita grammar 1.}
  \label{fig:g1_dfa}
\end{minipage}
\end{table*}


In the following experiments, we evaluate and compare different models for their rate of success in extracting the correct DFAs associated with the Tomita grammars. The results are presented in Figure~\ref{fig:g_bar_all}. Recall that the success rate of a recurrent model obtained with a grammar is calculated as the percentage of extracted DFAs with 100.0\% accuracy among all 130 DFAs. The ordinate labels the success rate in the range from 0.0\% to 100.0\%, and is increased by 10.0\% on the abscissa. The horizontal axis is labeled for each Tomita grammar with their corresponding index.

In Figure~\ref{fig:g_bar_all} we observe that the alignment between the overall variations of success rates and the changes of grammars' complexity is not as obvious as was previously shown in Figure~\ref{fig:acc_entropy}. This is as expected because when calculating the success rate, DFAs with close to 100.0\% accuracy are excluded. Recall from the results shown in Figure~\ref{fig:acc_entropy}, it can be seen that there are a considerable amount of ``good'' but not correct extracted DFAs. Despite this difference in the success rates obtained by different models across different grammars, we find that on grammars with lower complexity, all models are capable of producing correct DFAs. In particular, all models achieve much higher success rates on grammar 1. This may due to the reason that the DFA associated with grammar 1 has the fewest number of states (two states as shown in Figure~\ref{fig:g1_dfa}) and simplest state transitions among all other DFAs. Thus, the hidden vector space of all models is much easier to separate during training and identify during extraction. As for other grammars with lower complexity, their associated DFAs have both larger number of states and more complicated state transitions. Consider grammar 2 for example. While this grammar has the same level of complexity as grammar 1, its associated DFA has both a larger number of states and more complicated state transitions, as shown in Figure~\ref{fig:g2_dfa}. As such, the success rates obtained by most recurrent models on these grammars (except for grammar 1) rarely exceeds 50.0\%. 

Another interesting observation is that the experimental results shown in this and the previous sections both indicate that the performance of DFA extraction for second-order RNN is generally better than or comparable to that of other models on grammars with lower levels of complexity. For grammars with higher levels of complexity, the second-order RNN enables a much more accurate and stable DFA extraction. Also, generally speaking, MI-RNN provides the second best extraction performance. Recall that second-order RNN has a quadratic form of interaction between weights and neurons (introduced in Section~\ref{tab:rnn_models}). Thus, the multiplicative form of interaction used in MI-RNN can be regarded as an approximation to the quadratic interaction used in second-order RNN. This may imply that these special forms of interaction adopted by second-order RNN and MI-RNN are more suitable for generating spatially separable states and representing the state transition diagrams. Especially, we observe that on grammar 5 and 6, which have the highest complexity, only second-order RNN and MI-RNN are able to provide correct DFAs through extraction, while all other recurrent models fail. It is also worth noting that the Elman-RNN, especially Elman-RNN with the {\tt tanh} activation function,~\footnote{This result implies that the choice of activation function may also affect DFA extraction. A study of activation effects is a problem that could be included in future work.} obtains the worst success rates on most grammars. In particular, on grammar 2, while the accuracy of DFAs extracted from Elman-RNN is close to 100.0\% (as shown in Figure~\ref{fig:g2_line}), the success rate of Elman-RNN is only around 10\%. As for LSTM and GRU, their success rates are consistent with the results shown in Section~\ref{sec:acc}.

%% file: base/6_0_conclusion.tex
\section{Conclusion and Future Work}
We conducted a careful experimental study on learning and extracting deterministic finite state automata (DFA) from different recurrent networks, in particular the Elman-RNN, second-order RNN, MI-RNN, LSTM and GRU, from the Tomita grammars. We observe that the second-order RNN provides the best and most stable performance of DFA extraction on Tomita grammars in general. In particular, on certain grammars, the performance of DFA extraction for second-order RNN is significantly better than other recurrent models. Our experiments also show that, for all models except for second-order RNN, their performance of DFA extraction varies significantly across different Tomita grammars. This inconsistency is explained through our analysis on the complexity of Tomita grammars. Specifically, we introduce two metrics -- the entropy and average edit distance -- for describing the complexity of regular grammars with binary alphabet. Based on our metrics, we categorize seven Tomita grammars into three classes where each class has similar complexity. The categorization is consistent with the results observed in the experiments. 

We apply a generic compositional DFA extraction approach to all recurrent networks studied. Future work will include evaluating and comparing different DFA extraction approaches under the evaluation framework introduced in this study. Also, we could study and exploit the quadratic interaction taken by second-order RNN for training recurrent networks in order to combine desirable performance with reliable rule extraction. In addition, we intend to study the performance of DFA extraction on real-world applications. 
Another direction would be to see if other activation functions, such as ReLU can be used.
It would also be interesting to explore this approach on large scale grammar problems and to extend our theoretical analysis to more general grammars and those with larger alphabets.

\section{Acknowledgements}
We gratefully acknowledge partial support from the National Science Foundation.

%% file: base/appendix.tex
\appendices

\begin{figure*}[!t]
\centering
\begin{subfigure}{.22\textwidth}
  \centering
  \includegraphics[width=0.95\linewidth]{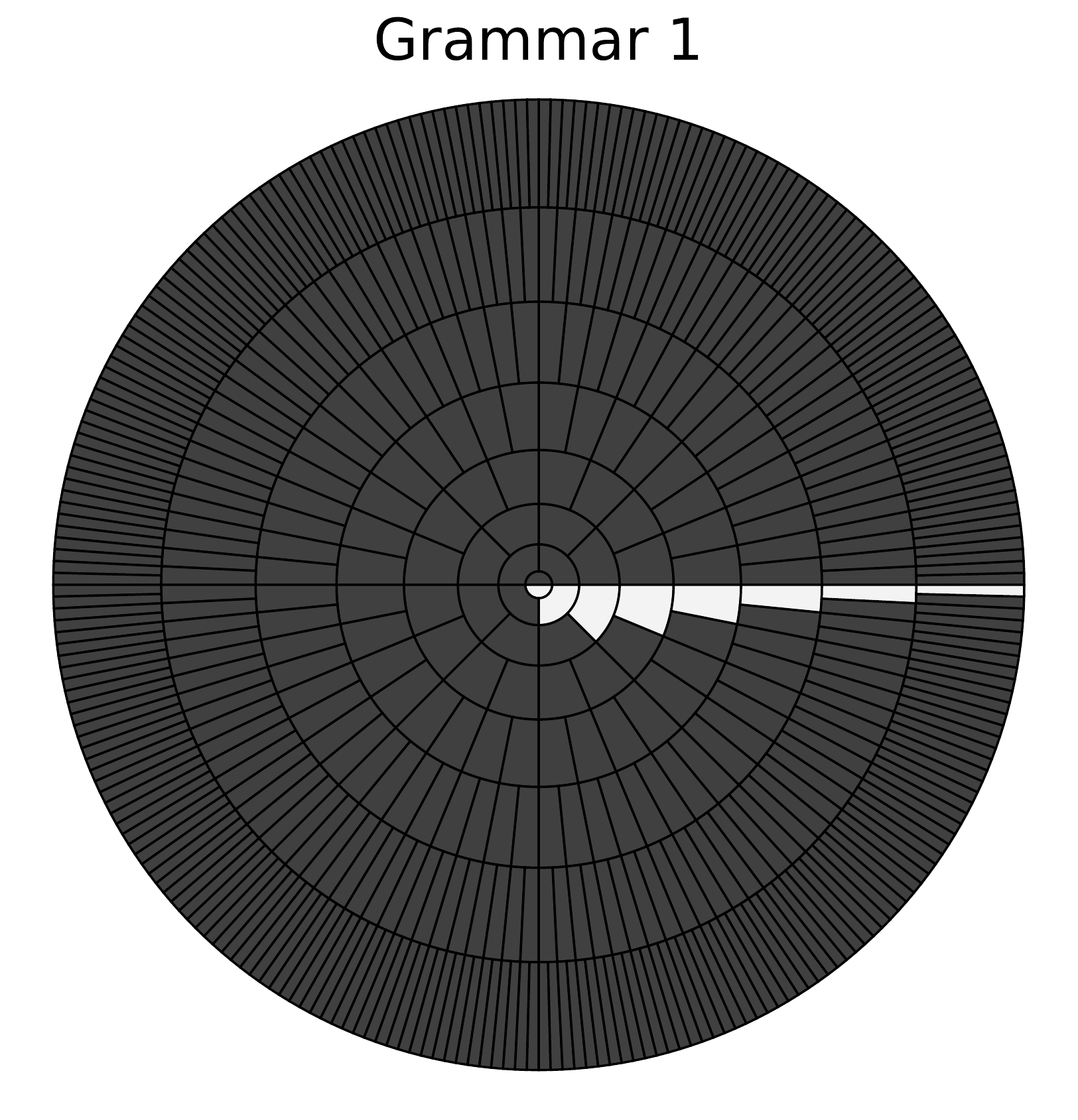}
  \vspace{-0.5em}
  \label{fig:g1_pie}
\end{subfigure} \hfill 
\begin{subfigure}{.22\textwidth}
  \centering
  \includegraphics[width=0.95\linewidth]{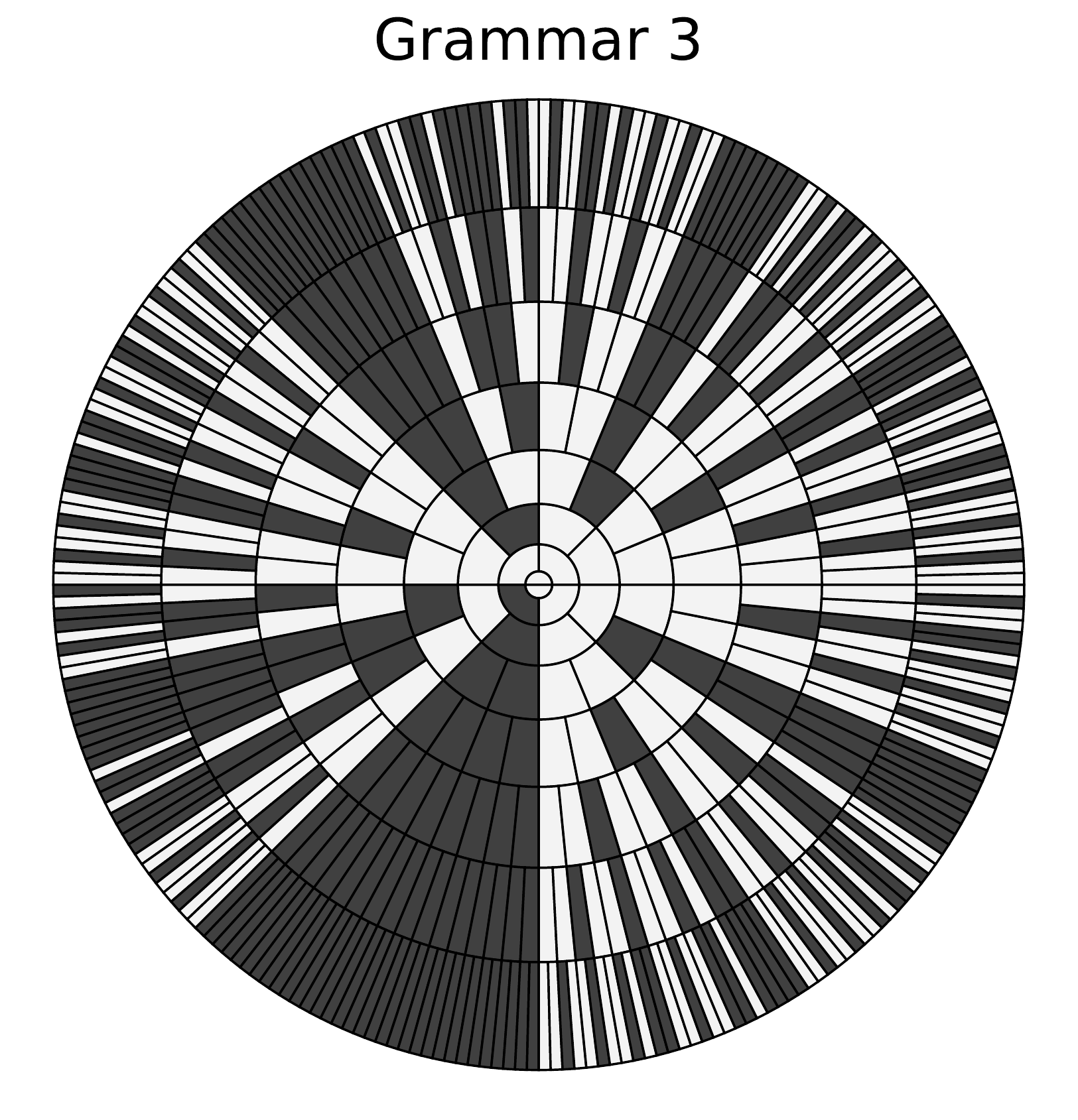}
  \vspace{-0.5em}
  \label{fig:g3_pie}
\end{subfigure} \hfill 
\begin{subfigure}{.22\textwidth}
  \centering
  \includegraphics[width=0.95\linewidth]{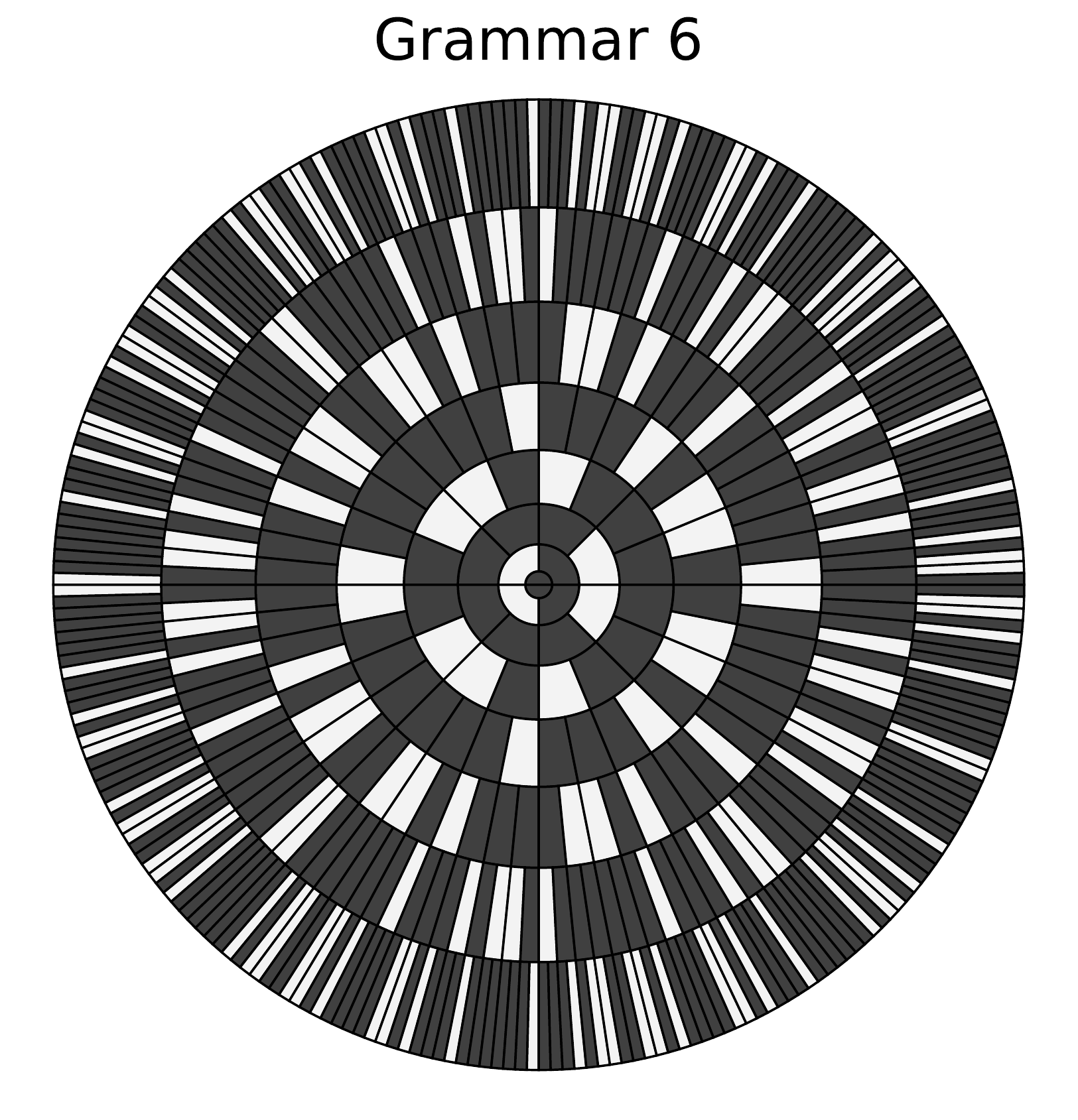}
  \vspace{-0.5em}
  \label{fig:g6_pie}
\end{subfigure} \hfill
\begin{subfigure}{.22\textwidth}
  \centering
  \includegraphics[width=0.95\linewidth]{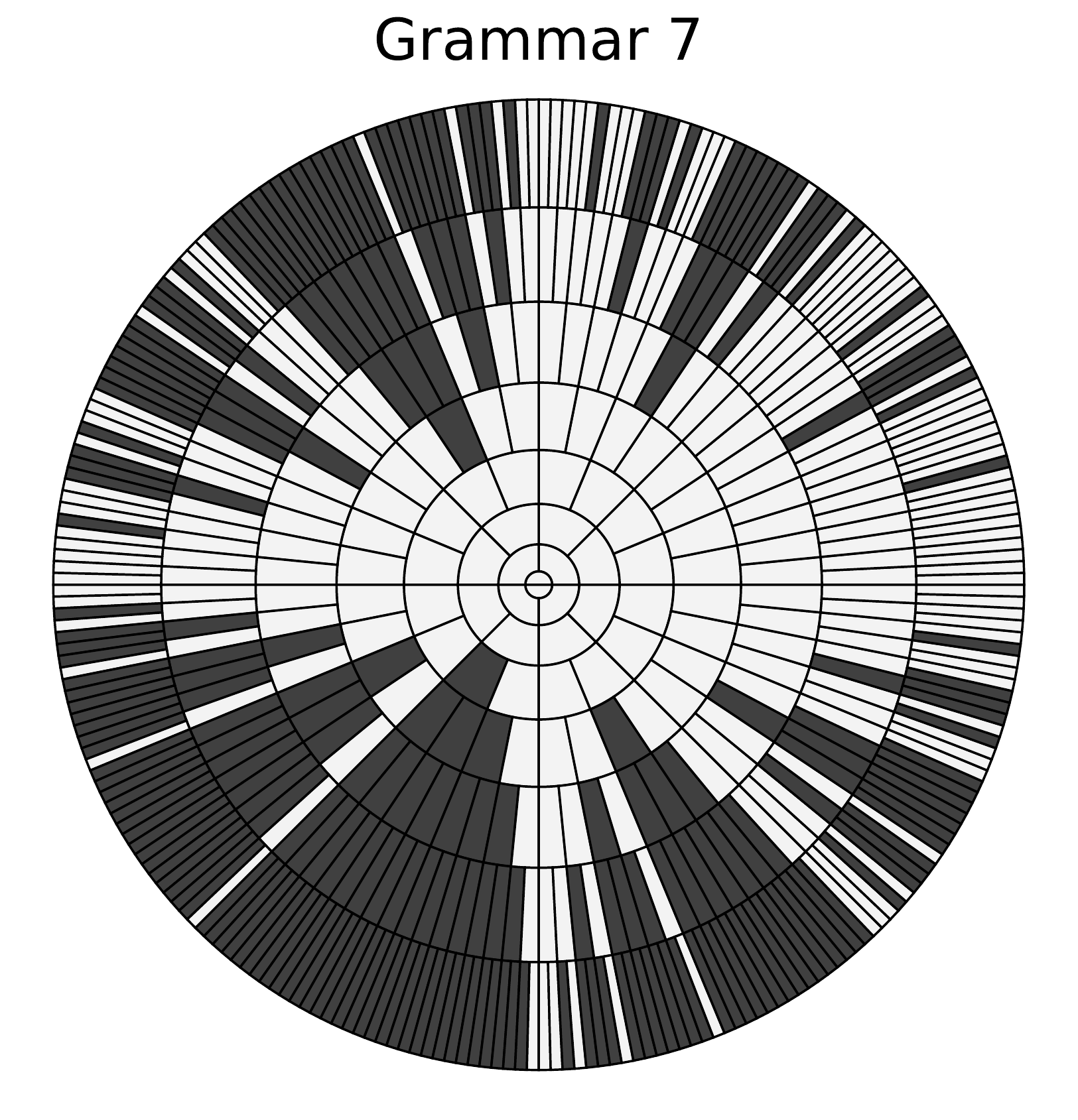}
  \vspace{-0.5em}
  \label{fig:g7_pie}
\end{subfigure} \hfill
\caption{Graphic representation of the distribution of strings ($1 \leq N \leq 8$) for grammars 1, 3, 6 and 7.}
\label{fig:pie_grammars_1367}
\end{figure*}

\begin{figure*}[!t]
\centering
\begin{subfigure}{.23\textwidth}
  \centering
  \includegraphics[width=\linewidth]{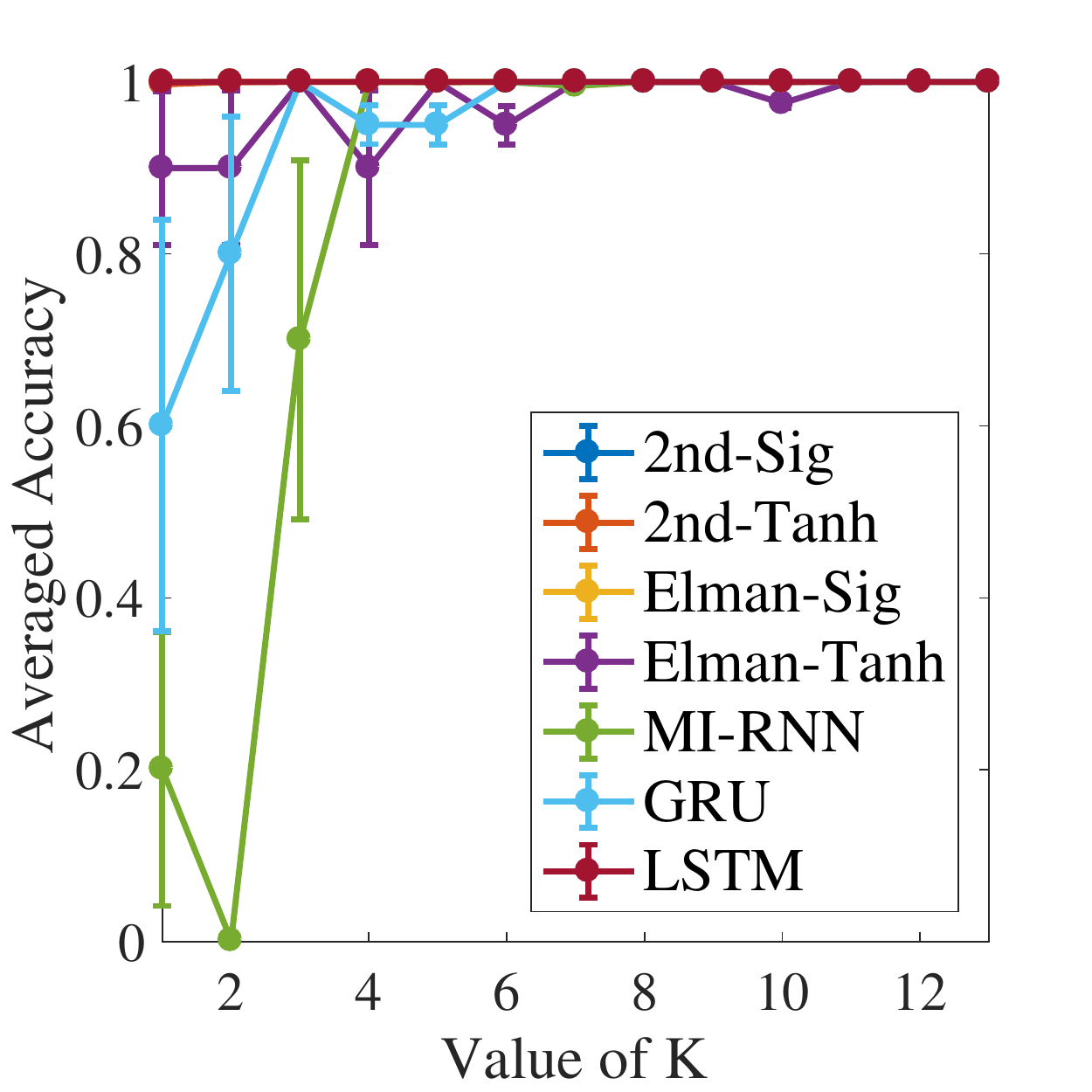}
  \centering
  \vspace{-0.5em}
  \caption{Grammar 1.}
  \label{fig:g1_line}
\end{subfigure} \hfill
\begin{subfigure}{.23\textwidth}
  \centering
  \includegraphics[width=\linewidth]{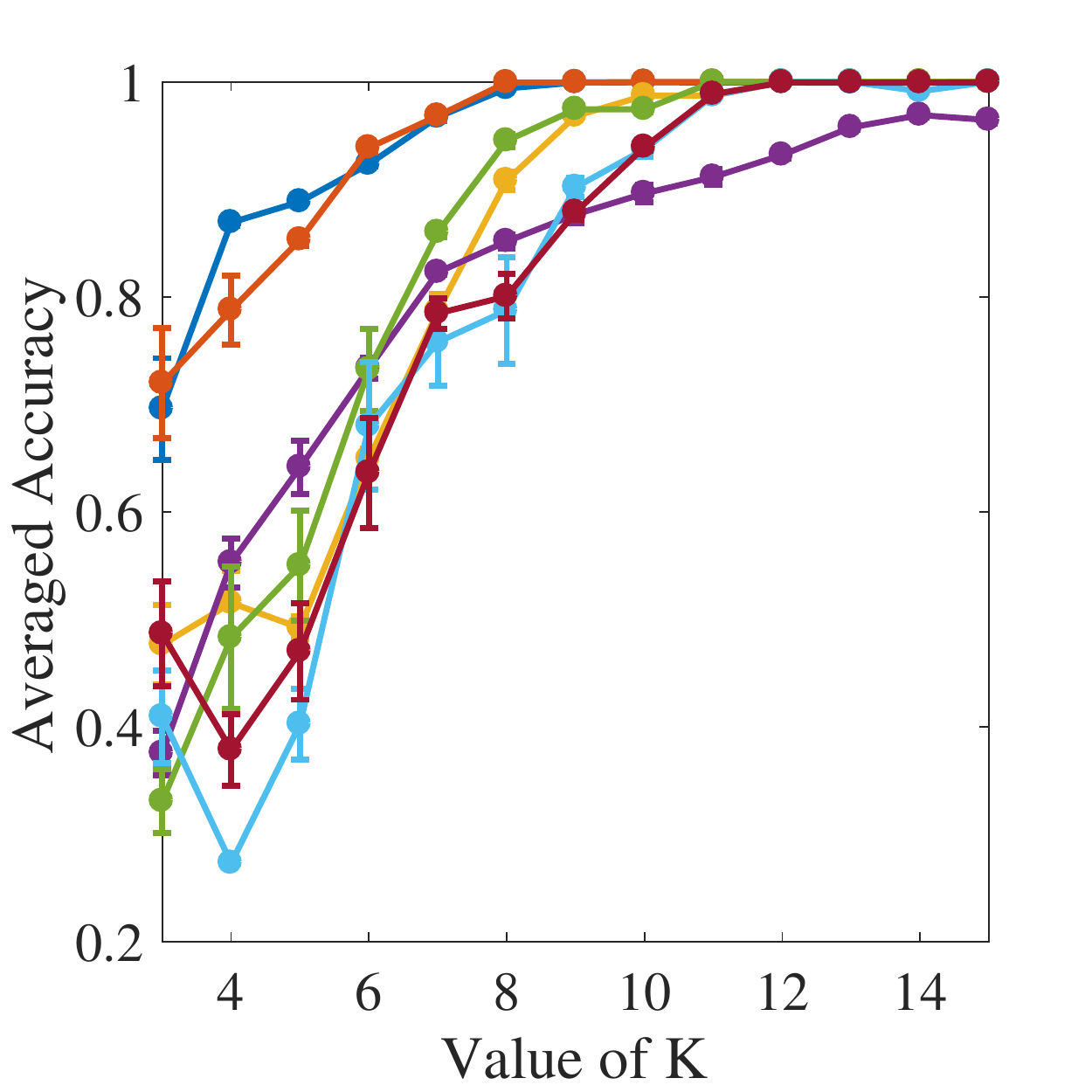}
  \vspace{-0.5em}
  \caption{Grammar 3.}
  \label{fig:g3_line}
\end{subfigure} \hfill 
\begin{subfigure}{.23\textwidth}
  \centering
  \includegraphics[width=\linewidth]{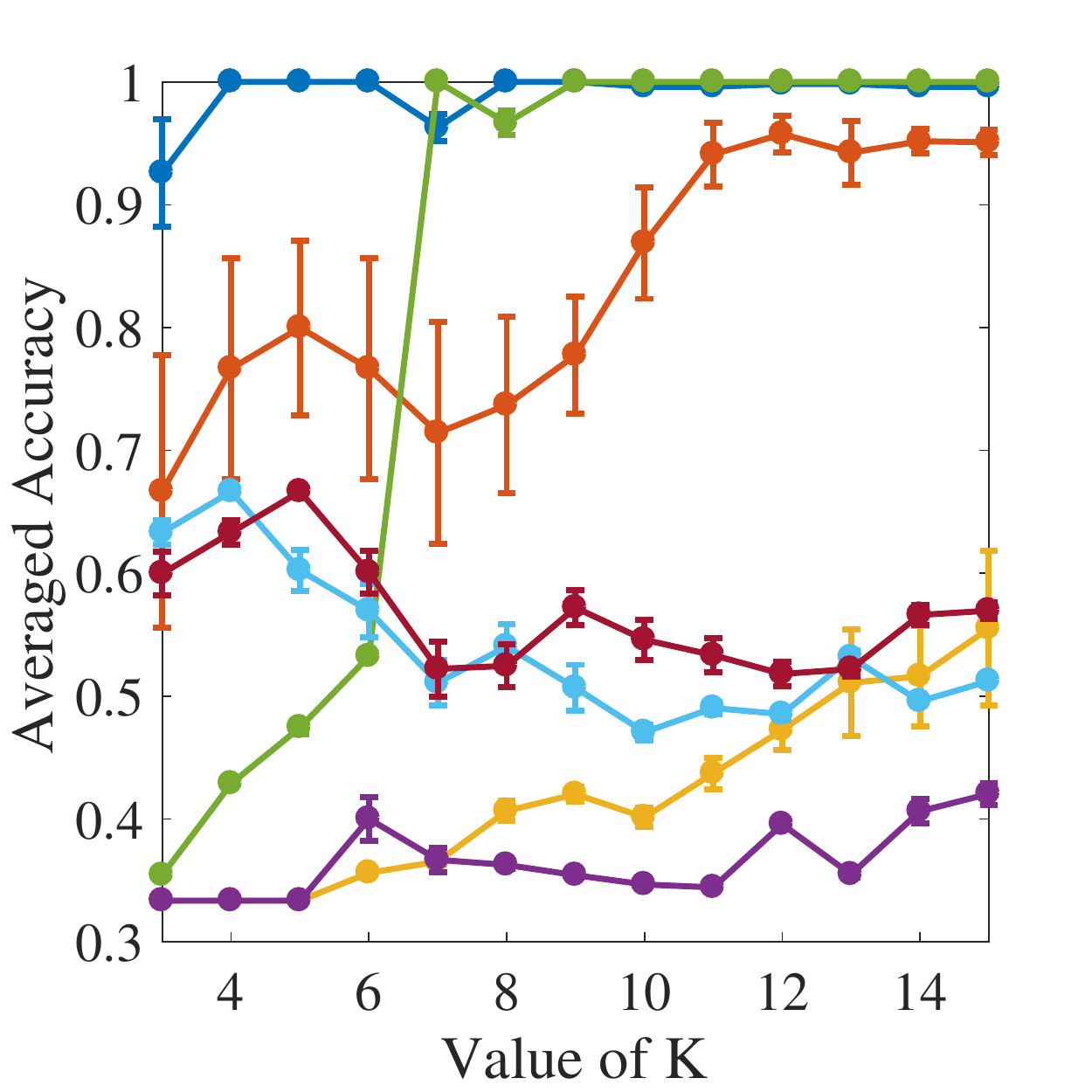}
  \vspace{-0.5em}
  \caption{Grammar 6.}
  \label{fig:g6_line}
\end{subfigure} \hfill
\begin{subfigure}{.23\textwidth}
  \centering
  \includegraphics[width=\linewidth]{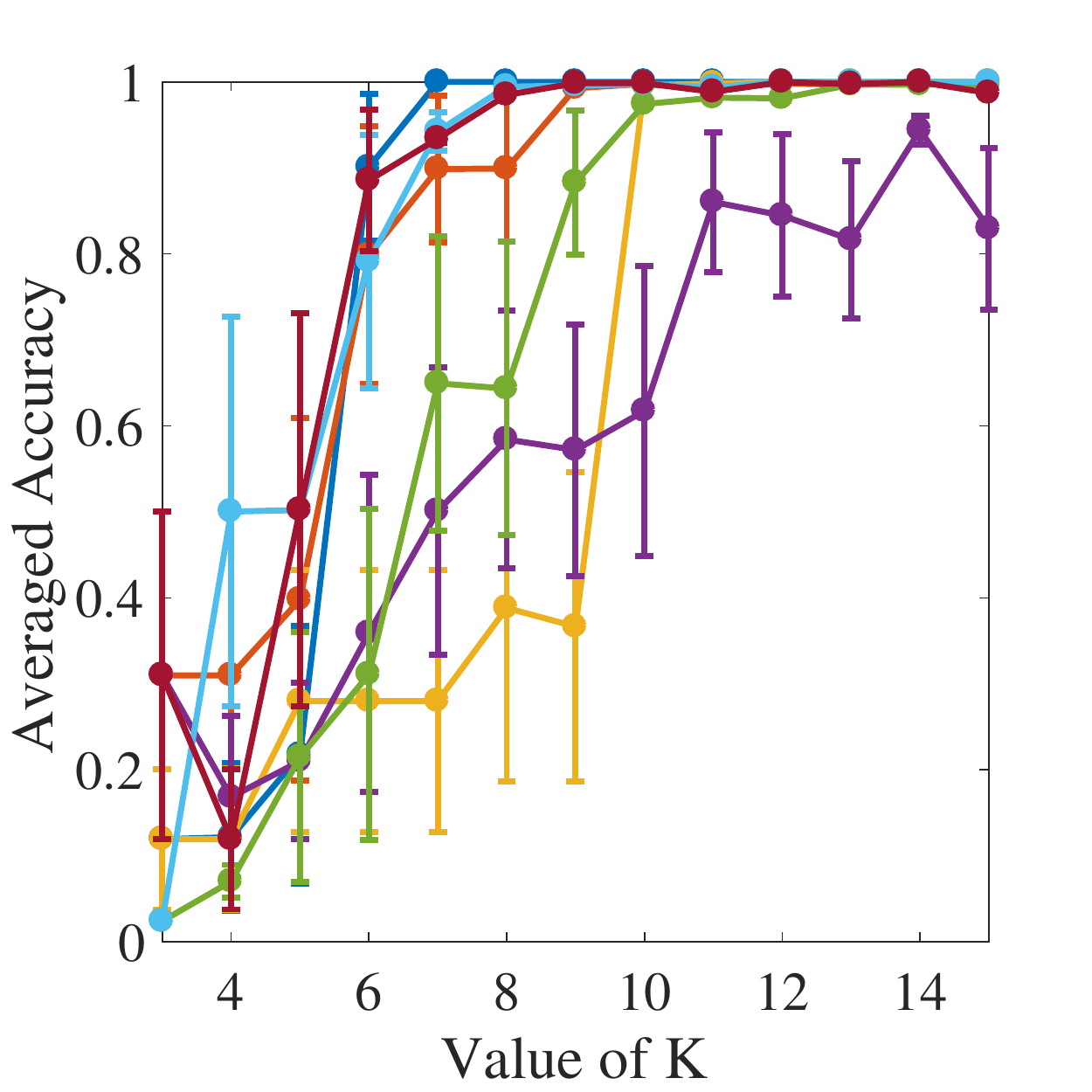}
  \vspace{-0.5em}
  \caption{Grammar 7.}
  \label{fig:g7_line}
\end{subfigure} \hfill
\caption{Mean and variance of the accuracy of DFAs extracted by all models on grammars 1, 3, 6 and 7.}
\label{fig:avg_var_1367}
\end{figure*}

\section{Proof of Proposition~\ref{prop:entropy}}
\label{sec:prop1_proof}
\begin{proof}
Given any concentric ring shown in Figure~\ref{fig:pie_grammars}, let $R$ denote the number of consecutive runs of strings and $R_P$ and $R_N$ denote the number of consecutive runs of positive strings and negative strings in this concentric ring respectively. Then we have $ \mathrm{E}[F]=\mathrm{E}[R] -1 = \mathrm{E}[R_p] + \mathrm{E}[R_n] - 1 $. Without loss of generality, we can choose the first position as $\theta = 0$ in the concentric ring. Then we introduce an indicator function $I$ by $I_i = 1$ representing that a run of positive strings starts at the $i$-th position and $I_i = 0$ otherwise. Since $ R_p = \sum_{i=1}^{2^N} I_i$, we have 
\begin{equation}
\vspace{-0.1em}
  \begin{aligned}
  \mathrm{E} [R_p] \!=\! \sum_{i=1}^{2^N} \mathrm{E} [I_i] \;\; \text{and} \;\; \mathrm{E} [I_i] \!=\! \begin{cases} m_{p}/{2^N}  , \qquad \qquad  \quad \; i = 1\\
 m_n m_P / 2^N (2^N-1) , \, i \neq 1. \nonumber
  \end{cases}
  \end{aligned}
  \vspace{-0.1em}
\label{eq:E_runs}
\end{equation}
As such, we have 
\begin{equation}
\vspace{-0.1em}
  \begin{aligned}
  \mathrm{E} [R_P] = \frac{m_{p}(1 + m_{n})}{2^N} \;\; \text{and} \;\; \mathrm{E}[R_N] = \frac{m_{n}(1 + m_{P})}{2^N}. \nonumber
  \end{aligned}
\label{eq:E_runs_p}
\end{equation}
By substituting $\mathrm{E}[F]$ into~\eqref{eq:entropy}, we have
\begin{equation}
  \begin{aligned}
  H(G) = 1 + \underset{N \rightarrow \infty }{\mathrm{lim\,sup}} \,\frac{\log_{2} \big(r_p(1 - r_p)\big)}{N}.
  \end{aligned}
\label{eq:entropy_calculation}
\end{equation}

\end{proof}

\section{Proof of Theorem~\ref{thm:entropy}}
\label{sec:appendix_entropy}
In both the Definition~\ref{def:entropy} and Proposition~\ref{prop:entropy}, we use $\limsup$  to cover certain particular cases, for instance when $N$ is set to odd value for grammar 5. In the following proof, without loss of generality, we use $\lim$ instead of $\limsup$ for simplicity. 

According to Proposition~\ref{prop:entropy}, for any regular grammar $G$ with binary alphabet, its entropy $H(G) \in [0,1]$. It can be checked that the maximum value of $H(G)$ is $1$ when $r_p = 0.5$. Also, the minimum value of $H(G)$ is $0$ and can be reached when $r_p = 0$ or $1$. However, $r_p = 0$ or $1$ are only allowed for grammars that either accept or reject any binary string, hence are not considered this theorem. As such, in our case, we take the value of entropy as minimum when $r_p = 1/2^N$ or $1 - 1/2^N$. In the following, we only discuss the former case due to space limit, the latter can be similarly derived. 

\begin{proof}
For each class of grammars, given that their $m_p$ takes the corresponding form shown in Theorem~\ref{thm:entropy}, the proof for the sufficient condition is trivial and can be checked by applying \emph{L'Hospital's Rule}. As such, in the following we only provide a proof for the necessary condition. 

From~\eqref{eq:entropy_calculation}, we have:
\begin{equation}
\vspace{-0.5em}
	\begin{aligned}
	\nonumber 
	H(G) &= \lim_{N \to \infty}\frac{\log_2(m_p \cdot 2^N-m_p^2)}{N}-1 \\
         &= \lim_{N \to \infty}\frac{m_p'\cdot 2^N+\ln2\cdot 2^N \cdot m_p-2m_p \cdot m_p'}{\ln2 \cdot (m_p \cdot 2^N-m_p^2)}-1 \\
         &= \lim_{N \to \infty}\frac{m_p' \cdot 2^N+\ln2\cdot m_p^2 - 2m_p \cdot m_p'}{\ln2 \cdot m_p \cdot (2^N-m_p)}.
	\end{aligned}
	\vspace{-0.5em}
\end{equation}

It is easy to check that $\lim_{N \to \infty}\frac{m_p'}{m_p}$ exists for regular grammars, then we separate the above equation as follows:
\begin{equation}
\vspace{-0.5em}
	\begin{aligned}
	\nonumber 
	H(G) = \lim_{N \to \infty}\frac{m_p'}{\ln2\cdot m_p} + 
	       \lim_{N \to \infty}\frac{1-\frac{m_p'}{\ln2\cdot m_p}}{\frac{2^N}{m_p}-1}.
    \end{aligned}
    \vspace{-0.1em}
\end{equation}
It should be noted that the second term in the above equation equals $0$. Specifically, assuming that $m_p$ has the form of $\alpha \cdot b^N$ where $b < 2$ ($b$ cannot be larger than 2 for binary alphabet), then the denominator of the second term is infinity. If $m_p$ has the form of $\alpha\cdot{2}^N$, then the numerator tends to zero while the denominator is finite. As such, we have
\begin{equation}
	\begin{aligned}
\nonumber H(G) = \lim_{N \to \infty}\frac{m_p'}{\ln2\cdot m_p}.
    \end{aligned}
    \vspace{-0.5em}
\end{equation}

If $H(G)=0$, then we have $\lim_{N \to \infty}\frac{m_p'}{m_p}=0$, indicating that the dominant part of $m_p$ has a polynomial form of $N$ hence $m_p \sim \mathrm{poly}(N)$. 

If $H(G)=t\neq 0$, then we have $\lim_{N \to \infty}\frac{\ln(m_p)}{tN\ln2}=1$, which gives that $m_p \sim  \beta \cdot{2}^{tN}$, where $\beta > 0$. If $t = \log_2 b$, then we have $m_p$ $\sim \beta \cdot{b}^{N}$ where $b < 2$. Furthermore, if $t=1$, we have $m_p \sim \alpha\cdot{2}^{N}$ where $\alpha \in [0,1)$.

\end{proof}

Here we calculate the $m_p$ for grammar 4, 5 and 7 which falls into each of the three classes of grammars, respectively. 
\begin{enumerate}[label=(\alph*)]
  \item $m_{p}(G_4)=\alpha \!\cdot\! b^N$, where $\alpha = 1/3\cdot(19+3\sqrt{33})^{1/3}+1/3\cdot(19-3\sqrt{33})^{1/3}+1/3$ and $\beta = \{3(586+102\sqrt{33})^{1/3} \}. /\{(586+102\sqrt{33})^{2/3}+4-2(586+102\sqrt{33})^{1/3}\}$. The calculation is similar to calculating \emph{Tribonacci number}~\cite{feinberg1963fibonacci}; 
  \item $m_{p}(G_5)=\alpha\cdot2^N$. When $N$ is odd/even, $\alpha=0$/$0.5$;
  \item We can classify all positive strings associated with grammar 7 into groups: $1^+0^+1^+0^+$, $0^+1^+0^+$, $1^+0^+1^+$, $1^+0^+$, $0^+1^+$, $1^+$ and $0^+$, where $^+$ indicates 1 or more repetitions. By simple combinatorics, we have $m_p(G_7)= C_{N-1}^3 + 2C_{N-1}^2 + 2C_{N-1}^{1} + 2$.
\end{enumerate}

\section{Calculation of the Average Edit Distance of Tomita Grammars}
\label{sec:appendix_dist}
\begin{enumerate}[label=(\alph*), wide, labelwidth=!, labelindent=0pt]
\item For grammar 1, 2 and 7, their corresponding $D(G_{1,2,7}) = \infty$ as $N \rightarrow \infty$. Take grammar 1 as an example. Given $N$, there is only one positive string $x_p$, which consists of $N$ 1's. According to~\eqref{eq:dist_neg_pos}, we have 
  \begin{equation}
  \label{eq:dist_neg_pos_g1}
    \begin{aligned}
    &\frac{1}{|X_{p}^N|} \sum_{x_p \in X_{p}^N} d_{edit}(x_p, X_{n}^N) = 1, \\ 
    &\frac{1}{|X_{n}^N|} \sum_{x_n \in X_{n}^N} d_{edit}(x_n, X_{p}^N) = \frac{1}{2^N - 1} (C_{N}^1 + \dots + NC_N^N) \\
    & \qquad \qquad \qquad \qquad \qquad \qquad = \frac{N}{2}\frac{2^N}{2^N - 1}.
    \end{aligned}
  \end{equation}
  By substituting~\eqref{eq:dist_neg_pos_g1} into~\eqref{eq:distance_g}, we have 
  \begin{equation}
  \begin{aligned}
  \label{eq:distance_g1}
  D(G_1) = \underset{N \rightarrow \infty}{\mathrm{lim}} \frac{1}{2} \big (\frac{N}{2}\frac{2^N}{2^N - 1}+1 \big ) = \infty.
  \end{aligned}
\end{equation}
Similar results can be obtained for grammar 2 and 7. 

\item For grammar 3 and 4, their average edit distance $D(G_{3,4}) > 1$. Take grammar 3 as an example, it is easy to check that $d_{edit}(x_p, X_{n}^N) = 1$ for any $x_p$. For a negative string $x_N = \dots vu \dots$ or $\dots uv \dots$, where $u$ denotes any substring that is recognized by grammar 3, and $v$ denotes any string that is rejected by grammar 3. The minimum $k$ substitutions required to convert $x_n$ into any $x_p$, depends on the number of occurrences of $v$, which can be larger than 1. As such, $\sum_{x_n \in X_{n}^N} d_{edit}(x_n, X_{p}^N) > |X_{n}^N|$, hence $D(G_{3}) > 1$.

\item For grammar 5 and 6, their average edit distance $D(G_{5, 6}) \equiv 1$. Specifically, for grammar 5, we only consider the case when $N$ is even, otherwise there are no positive strings hence $X_{p}^N$ is empty. Given that $N$ is even, it is clear that $d_{edit}(x_p, X_{n}^N) = d_{edit}(x_n, X_{p}^N)$ and $|X_{p}^N| = |X_{n}^N|$. Then we have $D(G_{5}) \equiv 1$. For grammar 6, it is easy to check that $d_{edit}(x_p, X_{n}^N) = 1$ for any $x_p$, and
\begin{equation}
  \begin{aligned}
  \label{eq:distance_g6}
  \sum_{x_n \in X_{n}^N} d_{edit}(x_n, X_{p}^N) = \begin{cases}
                                                  |X_{n}^N| + 2 & \text{ if } N  \equiv 1 \,\, (\mathrm{mod}\, 3) \\
                                                  |X_{n}^N|     & \text{ otherwise }.
                                                  \end{cases}
  \end{aligned}
\end{equation}
Then we have $D(G_{6}) = 1$ when $N \rightarrow \infty$.
\end{enumerate}

\section{Proof of Proposition~\ref{prop:relation}}
\label{sec:appendix_prop_2}
\begin{proof}

It is clear to show that 

\begin{equation}
  \begin{aligned}
  \label{eq:zero}
  0 \leq \limsup_{N \rightarrow \infty} \frac{1}{N} \log_2 D^{N}(G) \leq \limsup_{N \rightarrow \infty} \frac{1}{N} \log_2 N = 0. 
  \end{aligned}
\end{equation}

As such we have 
\begin{equation}
  \begin{aligned}
  \label{eq:distance_g_n}
 & \limsup_{N \rightarrow \infty}  \frac{1}{N} \log_2 (\overline{D}) \\
 & \; = \limsup_{N \rightarrow \infty} \frac{1}{N} \log_2 \big( \frac{r_p D_{n}^{N} + (1 - r_p) D_{p}^{N}}{2 r_p (1 - r_p) 2^N} \big) (2 r_p (1 - r_p) 2^N)  \\
 & \; \leq  \limsup_{N \rightarrow \infty} \frac{1}{N} \big (\log_2(2 r_p (1 - r_p) 2^N) + \log_2 D^{N}(G)\big ) \\
 & \; \leq  H(G) + \limsup_{N \rightarrow \infty} \frac{1}{N} \log_2 D^{N}(G).
  \end{aligned}
\end{equation}
where $\overline{D} = r_p D_{n}^{N} + (1 - r_p) D_{p}^{N}$. Since the sequence $\lim_{N \rightarrow \infty} \frac{1}{N} \log_2 D^{N}(G)$ is bounded and converges to zero, we have $H(G) = \limsup_{N \rightarrow \infty}  \frac{1}{N} \log_2 (\overline{D})$.
                               
\end{proof}

%% file: main.bbl
\begin{thebibliography}{10}
\providecommand{\url}[1]{#1}
\csname url@samestyle\endcsname
\providecommand{\newblock}{\relax}
\providecommand{\bibinfo}[2]{#2}
\providecommand{\BIBentrySTDinterwordspacing}{\spaceskip=0pt\relax}
\providecommand{\BIBentryALTinterwordstretchfactor}{4}
\providecommand{\BIBentryALTinterwordspacing}{\spaceskip=\fontdimen2\font plus
\BIBentryALTinterwordstretchfactor\fontdimen3\font minus
  \fontdimen4\font\relax}
\providecommand{\BIBforeignlanguage}[2]{{%
\expandafter\ifx\csname l@#1\endcsname\relax
\typeout{** WARNING: IEEEtran.bst: No hyphenation pattern has been}%
\typeout{** loaded for the language `#1'. Using the pattern for}%
\typeout{** the default language instead.}%
\else
\language=\csname l@#1\endcsname
\fi
#2}}
\providecommand{\BIBdecl}{\relax}
\BIBdecl

\bibitem{du2017topology}
J.~Du, S.~Zhang, G.~Wu, J.~M.~F. Moura, and S.~Kar, ``Topology adaptive graph
  convolutional networks,'' \emph{CoRR}, vol. abs/1710.10370, 2017.

\bibitem{omlin2000symbolic}
C.~W. Omlin and C.~L. Giles, ``Symbolic knowledge representation in recurrent
  neural networks: Insights from theoretical models of computation,''
  \emph{Knowledge based neurocomputing}, pp. 63--115, 2000.

\bibitem{minsky1967computation}
M.~L. Minsky, \emph{Computation: finite and infinite machines}.\hskip 1em plus
  0.5em minus 0.4em\relax Prentice-Hall, Inc., 1967.

\bibitem{BorgesGL11}
R.~V. Borges, A.~S. d'Avila Garcez, and L.~C. Lamb, ``Learning and representing
  temporal knowledge in recurrent networks,'' \emph{{IEEE} Trans. Neural
  Networks}, vol.~22, no.~12, pp. 2409--2421, 2011.

\bibitem{Lin96NARX}
T.~Lin, B.~G. Horne, P.~Ti{\~{n}}o, and C.~L. Giles, ``Learning long-term
  dependencies in {NARX} recurrent neural networks,'' \emph{{IEEE} Trans.
  Neural Networks}, vol.~7, no.~6, pp. 1329--1338, 1996.

\bibitem{dhingra2017linguistic}
B.~Dhingra, Z.~Yang, W.~W. Cohen, and R.~Salakhutdinov, ``Linguistic knowledge
  as memory for recurrent neural networks,'' \emph{CoRR}, vol. abs/1703.02620,
  2017.

\bibitem{murdoch2017automatic}
W.~J. Murdoch and A.~Szlam, ``Automatic rule extraction from long short term
  memory networks,'' \emph{CoRR}, vol. abs/1702.02540, 2017.

\bibitem{Empirical2017Wang}
Q.~Wang, K.~Zhang, A.~G. Ororbia~II, X.~Xing, X.~Liu, and C.~L. Giles, ``An
  empirical evaluation of rule extraction from recurrent neural networks,''
  \emph{Neural Computation}, vol.~30, no.~9, pp. 2568--2591, 2018.

\bibitem{weiss2017extracting}
G.~Weiss, Y.~Goldberg, and E.~Yahav, ``Extracting automata from recurrent
  neural networks using queries and counterexamples,'' \emph{Proceedings of the
  35th International Conference on Machine Learning, {ICML} 2018,
  Stockholmsm{\"{a}}ssan, Stockholm, Sweden, July 10-15, 2018}, pp. 5244--5253,
  2018.

\bibitem{cohen2017inducing}
M.~Cohen, A.~Caciularu, I.~Rejwan, and J.~Berant, ``Inducing regular grammars
  using recurrent neural networks,'' \emph{CoRR}, vol. abs/1710.10453, 2017.

\bibitem{omlin1996extraction}
C.~W. Omlin and C.~L. Giles, ``Extraction of rules from discrete-time recurrent
  neural networks,'' \emph{Neural Networks}, vol.~9, no.~1, pp. 41--52, 1996.

\bibitem{casey1996dynamics}
M.~Casey, ``The dynamics of discrete-time computation, with application to
  recurrent neural networks and finite state machine extraction,'' \emph{Neural
  computation}, vol.~8, no.~6, pp. 1135--1178, 1996.

\bibitem{giles1992learning}
C.~L. Giles, C.~B. Miller, D.~Chen, H.-H. Chen, G.-Z. Sun, and Y.-C. Lee,
  ``Learning and extracting finite state automata with second-order recurrent
  neural networks,'' \emph{Neural Computation}, vol.~4, no.~3, pp. 393--405,
  1992.

\bibitem{watrous1992induction}
R.~L. Watrous and G.~M. Kuhn, ``Induction of finite-state languages using
  second-order recurrent networks,'' \emph{Neural Computation}, vol.~4, pp.
  406--414, 1992.

\bibitem{jacobsson2005rule}
H.~Jacobsson, ``Rule extraction from recurrent neural networks: Ataxonomy and
  review,'' \emph{Neural Computation}, vol.~17, no.~6, pp. 1223--1263, 2005.

\bibitem{giles1991second}
C.~L. Giles, D.~Chen, C.~Miller, H.~Chen, G.~Sun, and Y.~Lee, ``Second-order
  recurrent neural networks for grammatical inference,'' in \emph{Neural
  Networks, 1991., IJCNN-91-Seattle International Joint Conference on},
  vol.~2.\hskip 1em plus 0.5em minus 0.4em\relax IEEE, 1991, pp. 273--281.

\bibitem{kolen1994fool}
J.~F. Kolen, ``Fool's gold: Extracting finite state machines from recurrent
  network dynamics,'' in \emph{Advances in neural information processing
  systems}, 1994, pp. 501--508.

\bibitem{elman1990finding}
J.~L. Elman, ``Finding structure in time,'' \emph{Cognitive science}, vol.~14,
  no.~2, pp. 179--211, 1990.

\bibitem{hochreiter1997long}
S.~Hochreiter and J.~Schmidhuber, ``Long short-term memory,'' \emph{Neural
  computation}, vol.~9, no.~8, pp. 1735--1780, 1997.

\bibitem{cho2014properties}
K.~Cho, B.~Van~Merri{\"e}nboer, D.~Bahdanau, and Y.~Bengio, ``On the properties
  of neural machine translation: Encoder-decoder approaches,'' in
  \emph{Proceedings of SSST@EMNLP 2014, Eighth Workshop on Syntax, Semantics
  and Structure in Statistical Translation, Doha, Qatar, 25 October 2014},
  2014, pp. 103--111.

\bibitem{wu2016multiplicative}
Y.~Wu, S.~Zhang, Y.~Zhang, Y.~Bengio, and R.~Salakhutdinov, ``On multiplicative
  integration with recurrent neural networks,'' in \emph{Advances in Neural
  Information Processing Systems 29: Annual Conference on Neural Information
  Processing Systems 2016, December 5-10, 2016, Barcelona, Spain}, 2016, pp.
  2856--2864.

\bibitem{li2016kernel}
K.~Li and J.~C. Pr{\'\i}ncipe, ``The kernel adaptive
  autoregressive-moving-average algorithm,'' \emph{IEEE transactions on neural
  networks and learning systems}, vol.~27, no.~2, pp. 334--346, 2016.

\bibitem{tomita1982}
M.~Tomita, ``Dynamic construction of finite automata from example using
  hill-climbing,'' \emph{Proceedings of the Fourth Annual Cognitive Science
  Conference}, pp. 105--108, 1982.

\bibitem{chomsky1956three}
N.~Chomsky, ``Three models for the description of language,'' \emph{IRE
  Transactions on information theory}, vol.~2, no.~3, pp. 113--124, 1956.

\bibitem{du2016convergence}
J.~Du, S.~Ma, Y.~Wu, S.~Kar, and J.~M.~F. Moura, ``Convergence analysis of
  distributed inference with vector-valued gaussian belief propagation,''
  \emph{Journal of Machine Learning Research}, vol.~18, pp. 172:1--172:38,
  2017.

\bibitem{hopcroft2006automata}
J.~E. Hopcroft, R.~Motwani, and J.~D. Ullman, ``Introduction to automata
  theory, languages, and computation - international edition {(2.} ed),'' 2003.

\bibitem{giles1990higher}
C.~L. Giles, G.-Z. Sun, H.-H. Chen, Y.-C. Lee, and D.~Chen, ``Higher order
  recurrent networks and grammatical inference,'' in \emph{Advances in neural
  information processing systems}, 1990, pp. 380--387.

\bibitem{Murdoch18Beyond}
W.~J. Murdoch, P.~J. Liu, and B.~Yu, ``Beyond word importance: Contextual
  decomposition to extract interactions from lstms,'' \emph{CoRR}, vol.
  abs/1801.05453, 2018.

\bibitem{Murdoch18Hierarchical}
C.~Singh, W.~J. Murdoch, and B.~Yu, ``Hierarchical interpretations for neural
  network predictions,'' \emph{CoRR}, vol. abs/1806.05337, 2018.

\bibitem{pmlr-v80-weiss18a}
G.~Weiss, Y.~Goldberg, and E.~Yahav, ``Extracting automata from recurrent
  neural networks using queries and counterexamples,'' in \emph{Proceedings of
  the 35th International Conference on Machine Learning}, ser. Proceedings of
  Machine Learning Research, vol.~80.\hskip 1em plus 0.5em minus 0.4em\relax
  PMLR, 10--15 Jul 2018, pp. 5247--5256.

\bibitem{Angluin87}
D.~Angluin, ``Learning regular sets from queries and counterexamples,''
  \emph{Inf. Comput.}, vol.~75, no.~2, pp. 87--106, 1987.

\bibitem{gori1998inductive}
M.~Gori, M.~Maggini, E.~Martinelli, and G.~Soda, ``Inductive inference from
  noisy examples using the hybrid finite state filter,'' \emph{IEEE
  Transactions on Neural Networks}, vol.~9, no.~3, pp. 571--575, 1998.

\bibitem{zeng1993learning}
Z.~Zeng, R.~M. Goodman, and P.~Smyth, ``Learning finite state machines with
  self-clustering recurrent networks,'' \emph{Neural Computation}, vol.~5,
  no.~6, pp. 976--990, 1993.

\bibitem{Ribeiro0G16}
M.~T. Ribeiro, S.~Singh, and C.~Guestrin, ``"why should {I} trust you?":
  Explaining the predictions of any classifier,'' in \emph{Proceedings of the
  22nd {ACM} {SIGKDD} International Conference on Knowledge Discovery and Data
  Mining, San Francisco, CA, USA, August 13-17, 2016}, 2016, pp. 1135--1144.

\bibitem{Hinton17Distilling}
N.~Frosst and G.~E. Hinton, ``Distilling a neural network into a soft decision
  tree,'' in \emph{Proceedings of the First International Workshop on
  Comprehensibility and Explanation in {AI} and {ML} 2017 co-located with 16th
  International Conference of the Italian Association for Artificial
  Intelligence (AI*IA 2017), Bari, Italy, November 16th and 17th, 2017.}, 2017.

\bibitem{carrasco1997accurate}
R.~C. Carrasco, ``Accurate computation of the relative entropy between
  stochastic regular grammars,'' \emph{RAIRO-Theoretical Informatics and
  Applications}, vol.~31, no.~5, pp. 437--444, 1997.

\bibitem{thollard2000probabilistic}
F.~Thollard, P.~Dupont, C.~de~la Higuera \emph{et~al.}, ``Probabilistic dfa
  inference using kullback-leibler divergence and minimality,'' in \emph{ICML},
  2000, pp. 975--982.

\bibitem{lind1995introduction}
D.~Lind and B.~Marcus, \emph{An introduction to symbolic dynamics and
  coding}.\hskip 1em plus 0.5em minus 0.4em\relax Cambridge university press,
  1995.

\bibitem{de2010grammatical}
C.~De~la Higuera, \emph{Grammatical inference: learning automata and
  grammars}.\hskip 1em plus 0.5em minus 0.4em\relax Cambridge University Press,
  2010.

\bibitem{tieleman2012lecture}
T.~Tieleman and G.~Hinton, ``Lecture 6.5-rmsprop: Divide the gradient by a
  running average of its recent magnitude,'' \emph{COURSERA: Neural networks
  for machine learning}, vol.~4, no.~2, pp. 26--31, 2012.

\bibitem{frasconi1996representation}
P.~Frasconi, M.~Gori, M.~Maggini, and G.~Soda, ``Representation of finite state
  automata in recurrent radial basis function networks,'' \emph{Machine
  Learning}, vol.~23, no.~1, pp. 5--32, 1996.

\bibitem{sanfeliu1994active}
A.~Sanfeliu and R.~Alquezar, ``Active grammatical inference: a new learning
  methodology,'' in \emph{in Shape, Structure and Pattern Recogniton, D. Dori
  and A. Bruckstein (eds.), World Scientific Pub}.\hskip 1em plus 0.5em minus
  0.4em\relax Citeseer, 1994.

\bibitem{tivno1995learning}
P.~Ti{\v{n}}o and J.~{\v{S}}ajda, ``Learning and extracting initial mealy
  automata with a modular neural network model,'' \emph{Neural Computation},
  vol.~7, no.~4, pp. 822--844, 1995.

\bibitem{schellhammer1998knowledge}
I.~Schellhammer, J.~Diederich, M.~Towsey, and C.~Brugman, ``Knowledge
  extraction and recurrent neural networks: An analysis of an elman network
  trained on a natural language learning task,'' in \emph{Proceedings of the
  Joint Conferences on New Methods in Language Processing and Computational
  Natural Language Learning}.\hskip 1em plus 0.5em minus 0.4em\relax
  Association for Computational Linguistics, 1998, pp. 73--78.

\bibitem{feinberg1963fibonacci}
M.~Feinberg, ``Fibonacci-tribonacci,'' \emph{The Fibonacci Quarterly}, vol.~1,
  no.~1, pp. 71--74, 1963.

\end{thebibliography}
